\documentclass{article}
\usepackage{arxiv,times}

\usepackage[T1]{fontenc}
\usepackage[utf8]{inputenc}
\usepackage{amsmath,amssymb}
\usepackage{graphicx}
\usepackage{xcolor}
\usepackage{booktabs,array,tabularx,colortbl}
\usepackage{enumitem}
\usepackage{float}
\usepackage{pifont}
\usepackage[most]{tcolorbox}
\usepackage[colorlinks=true,citecolor=teal,linkcolor=teal,urlcolor=teal]{hyperref}
\usepackage{capt-of}
\newcommand{\cmark}{\ding{51}}
\newcommand{\xmark}{\ding{55}}
\newcommand{\notapplicable}{\textnormal{n/a}}
\newcolumntype{Y}{>{\raggedright\arraybackslash}X}

\newtcolorbox{promptbox}[1]{enhanced,breakable,colback=gray!4,colframe=gray!50,boxrule=0.4pt,arc=2pt,
  left=5pt,right=5pt,top=3pt,bottom=3pt,toptitle=1.5pt,bottomtitle=1.5pt,colbacktitle=gray!16,coltitle=black,
  fonttitle=\sffamily\bfseries\scriptsize,fontupper=\scriptsize,title={#1},before skip=4pt,after skip=6pt}
\newcommand{\ph}[1]{\textcolor{teal!65!black}{\textit{$\langle$#1$\rangle$}}}
\newcommand{\pc}[1]{\texttt{\detokenize{#1}}}

\floatstyle{plaintop}
\newfloat{algorithm}{t}{loa}
\floatname{algorithm}{Algorithm}
\newlist{pseudo}{enumerate}{1}
\setlist[pseudo]{label=\scriptsize\arabic*{:}, ref=\arabic*, leftmargin=2.3em,
                 labelsep=0.5em, topsep=2pt, itemsep=0.5pt, parsep=0pt, font=\normalfont}
\newcommand{\algind}{\hspace*{1.1em}}
\newcommand{\algkw}[1]{\textbf{#1}}
\newenvironment{algbox}{\vspace{0.4ex}\hrule\vspace{0.6ex}}{\vspace{0.4ex}\hrule\vspace{0.4ex}}

\newcommand{\projecturl}{\url{https://sirui-xu.github.io/InterEvolve}}
\title{InterEvolve: Test-Time Evolution of Reward Programs for Humanoid Loco-Manipulation}
\author{Zhuo Lin$^{\dag}$ \quad Sirui Xu$^{\dag}$ \quad Liuyu Bian \quad Yu-Xiong Wang$^{\ddag}$ \quad Liang-Yan Gui$^{\ddag}$\\
\normalfont\small{University of Illinois Urbana-Champaign}\\
\normalfont\small{$^{\dag}$Equal Contribution \quad $^{\ddag}$Equal Advising}\\
\normalfont\small\projecturl}

\begin{document}
\maketitle

\begin{center}
  \centering
  \vspace{-3em}
  \includegraphics[width=\linewidth]{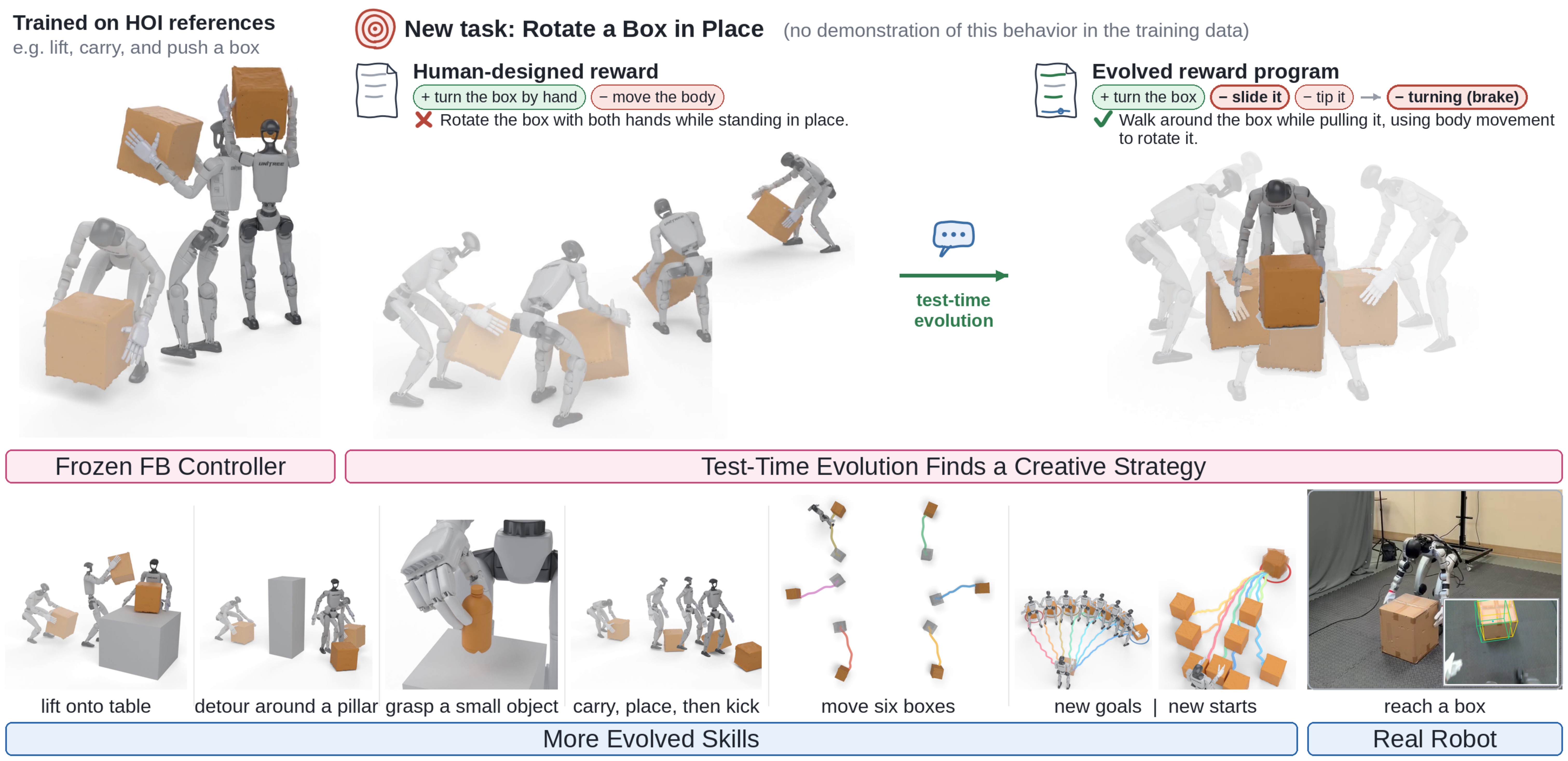}
  \vspace{-2em}
  \captionof{figure}{\textbf{InterEvolve adapts a frozen whole-body controller with novel use.} The controller trained on human-object interaction (HOI), such as lifting and pushing a box, is asked to rotate a box around in place, a behavior absent from its training data (Sec.~\ref{sec:app:coverage} and Figure~\ref{fig:umap}). A human-designed reward fails. The reward program that InterEvolve evolves compensate with novel body used to succeed. \textbf{Bottom:} InterEvolve can achieve versatile evolved skills in simulation and on a real robot. More demos are on the \href{https://sirui-xu.github.io/InterEvolve}{project page}.}
    
  \label{fig:loop}
\end{center}
\begin{abstract}
We study test-time evolution for humanoid loco-manipulation: solving tasks that a controller was never trained for by repurposing its existing skills, improving from its own attempts, and retaining what it learns, without retraining. Our key insight is that \textit{a broad controller already holds much of the competence a new task needs}, and that this competence becomes accessible through an interface between planning and control that is expressive enough to specify contact-rich, multi-stage interactions, yet executable and measurable enough that execution feedback can guide planning from experience. InterEvolve realizes this interface with two components. First, we develop an object-aware forward-backward (FB) behavioral foundation model, whose object residuals on a frozen body prior turn a new reward about the body or objects into loco-manipulation behavior at test time. Second, we specify tasks as \textit{reward programs}: staged rewards with completion conditions and tunable constants. A large language model (LLM) agent revises the program structure in context, drawing on execution feedback and a skill library of verified programs, while a numerical optimizer tunes its constants. With every candidate verified across parallel simulation scenarios, the program explores new ways to induce, repurpose, and compose the controller's existing motor competence for the task at hand, and thus improves over iterations. Experiments show that human-designed rewards leave much of the FB model's loco-manipulation competence untapped, whereas the programs InterEvolve evolves release it, sometimes through novel strategies. It further produces behaviors for diverse tasks, complex scenes, and long-horizon compositions in simulation, and evolved skills run autonomously on a physical Unitree G1 from egocentric onboard perception.

\end{abstract}

\section{Introduction}
\label{sec:intro}

Learned humanoid controllers now provide broad whole-body motion for real-robot control~\citep{sonic2025}, as well as contact-rich loco-manipulation~\citep{he2024omnih2o}, trained from human data. Yet a humanoid in an open-ended environment will meet tasks and scenes that were not enumerated during training. Consider a controller that has learned to push, lift, and carry boxes from human demonstrations. When asked to tip a box onto another face, it does not have a demonstration of this behavior, although the motions it needs lie within what it can already do. We study \textit{test-time evolution}, in which a humanoid finds such uses of its existing skills, improves from its own attempts within a budget of trials, and keeps what it learns, so that a later long-horizon composite task, such as carrying a box, placing it, and then kicking it, or a novel box tip, can build on earlier experience. What evolves is the task strategies, meaning which objectives to use or what stages to decompose, together with the experience retained across tasks, while the controller stays fixed.

Achieving this capability requires a way to state a new task that is expressive enough for contact-rich, multi-stage interaction, yet cheap to revise after each attempt. Common task interfaces for humanoid control, such as motion references~\citep{yang2025omniretarget}, goal states~\citep{he2026ultra}, and skill labels~\citep{wang2026humanx}, request desired motions, target states, or predefined behaviors. An objective such as tipping an object must therefore be translated into a reference goal or a new skill, either of which can be difficult to design for nuanced interactions. Rewards offer a more expressive alternative: they can specify contact-rich objectives and indicate where an attempt fell short, as their role in shaping such behavior during training demonstrates~\citep{andrychowicz2020learning}. Conventionally, however, each reward revision requires training a new policy, making it slow to iterate over many candidates~\citep{ma2024eureka}.

We introduce \textit{InterEvolve}, which builds this interface from reward programs and a behavioral foundation model that executes them without retraining (Fig.~\ref{fig:loop}). A \textit{reward program} is a sequence of stages, each with a reward and a completion condition, plus tunable weights and thresholds. To execute a program, we use a forward-backward (FB) behavioral foundation model~\citep{touati2021fb,tirinzoni2025motivo,li2026bfmzero}, which maps a reward to a latent that elicits the corresponding behavior from one fixed policy. Existing humanoid FB models, however, observe only the body, so rewards that differ only in the object's goal can collapse into one behavior. We extend the model in two ways: we attach trainable object residuals that read object features to its frozen body networks, and we train these residuals on large-scale human-object interaction (HOI) data.

With execution in place, the central challenge becomes reward formulation itself. It is especially pronounced in loco-manipulation, where a naively handcrafted reward underperforms even if the controller contains the relevant motor capabilities. InterEvolve therefore lets a large language model (LLM) agent iterates reward programs \textit{in context}: with its weights fixed, the agent combines task intent and execution feedback from simulation to refine the rewards and their composition. Instead of collecting references and training a policy for each new behavior, InterEvolve spends compute proposing, tuning, and verifying reward programs. This mirrors test-time scaling in language models, where more inference compute yields better solutions~\citep{snell2025scaling,brown2024large}.

We organize this evolution around two complementary objectives: improving the reward program for the current task and retaining what is learned to guide future tasks. (\textbf{i}) Two parts of a program must be searched together: its \textit{structure}, meaning which rewards and stages it uses, and its \textit{constants}, meaning the weights and thresholds. A promising structure can still fail under poor weights, so the agent revises the structure while an inner search with the covariance matrix adaptation evolution strategy (CMA-ES)~\citep{hansen2001cmaes} tunes the constants of each proposal. Parallel simulation supplies batches of execution evidence, so each revision is guided by a comparison over many rollouts. (\textbf{ii}) For future tasks, InterEvolve summarizes successful reward programs as a \textit{skill library}. Later tasks can retrieve these experiences without rediscovering them from scratch.

Our contributions are threefold. First, a framework for self-evolving humanoid loco-manipulation that shifts task-specific compute from training to test-time evolution: an LLM agent learns in context to write and adapt reward programs that a fixed, reusable controller executes for new tasks and scenes. Second, an object-aware FB behavioral foundation model that translates reward objectives into whole-body interaction. Third, an evaluation of how reward design and accumulated experience affect reference tracking and goal-conditioned tasks, showing that success grows with evolution. We further show unseen tasks and long-horizon compositions in simulation and fully autonomous deployment on a physical Unitree G1 from onboard perception.

\section{Related Work}
\label{sec:related}

\noindent\textbf{Humanoid loco-manipulation and task interfaces.}
Humanoid control has grown reusable, from simulated characters \citep{deepmimic,peng2022ase} to real-robot whole-body policies learned from teleoperation and human motion \citep{fu2024humanplus,he2024omnih2o,ji2024exbody2,chen2025gmt,yin2026unitracker,liao2025beyondmimic,ze2025twist,sonic2025}. Loco-manipulation policies learn grasping and contact-rich interaction from references \citep{luo2024omnigrasp,xu2025intermimic,tessler2025maskedmanipulator,wang2025skillmimic,xu2026interprior} and transfer to real robots \citep{liu2024opt2skill,li2024okami,sun2025ulc,yang2025omniretarget,zhao2025resmimic,fu2025demohlm,wang2025physhsi,he2026ultra}. Others condition control on multimodal prompts \citep{kareer2024egomimic,xue2025leverb,ding2025humanoidvla,deng2025human,kalaria2025dreamcontrol,jiang2026wholebodyvla,xie2026grail}, sequences existing skills with planners~\citep{yuan2025being0,wen2025coa,ren2026cybowaiter,sun2026choreo,xiao2023unihsi,tevet2024closd}, or imitates video-imagined interactions~\citep{chen2026imagine2real}. InterEvolve instead states the task as a reward program, which says which contacts and stage transitions matter, and revises it after each attempt.

\noindent\textbf{Behavioral foundation models and reward inference.}
Successor features decouple occupancy from reward \citep{dayan1993successor,barreto2017successor}, and FB representations factorize the successor measure so a backward map projects any reward into a latent prompt \citep{touati2021fb,touati2023zeroshot}. Humanoid behavioral foundation models thus run one policy for new rewards, goals, or references without task-specific training \citep{tirinzoni2025motivo,li2026bfmzero}. Follow-ups refines the representation \citep{cetin2025finer,bagatella2026tdjepa}, searches the latent space \citep{sikchi2025fastadapt}, infers tasks online \citep{rupf2025opti,bagot2026exploration}, grounds language \citep{sikchi2025rlzero}. These models observe only the body, so rewards that differ only in what happens to an object collapse into one behavior. Our object-aware FB model removes this limit.

\noindent\textbf{Reward design and agents that learn from execution.}
Language models write reward functions from task descriptions \citep{xie2024text2reward}, refine them from simulator feedback \citep{ma2024eureka,ma2024dreureka}, and search over reward designs \citep{zhang2025orso,gao2025rfagent,lee2026rda}, including for humanoid locomotion \citep{wu2025stride}, but each candidate reward is realized by training a policy. Language to Rewards optimizes LLM-written rewards with model predictive control instead \citep{yu2023l2r,liang2024lmpc}. MotionDisco evolves humanoid motions with an LLM and trajectory optimization before training trackers for them \citep{taouil2026motiondisco}, and ROSETTA builds multi-stage reward programs from language preferences \citep{srivastava2026rosetta}. Coding agents likewise refine executable plans from feedback and retain reusable experience \citep{liang2023codeaspolicies,zha2024droc,zhou2024soar,nvidia2026aspire,elmaaroufi2026rho,xiao2026enpire,wang2026shaper}. In InterEvolve, each candidate reward costs a batch of rollouts on a fixed controller rather than an RL run, and verified programs form a skill library for later tasks.

\noindent\textbf{Test-time compute and search.}
Language models improve with more inference compute through repeated sampling, step-level search, and long reasoning~\citep{snell2025scaling,brown2024large,guo2025deepseekr1}, and program-search systems evolve code against an automatic evaluator~\citep{romeraparedes2024funsearch,novikov2025alphaevolve}. In control, test-time search usually plans actions with a dynamics model~\citep{hansen2024tdmpc2} or steers a pretrained humanoid policy~\citep{zhang2026sumo,seo2026rgb,cao2026texedo}. InterEvolve searches over task specifications instead, keeping the controller fixed and verifying proposed reward programs in parallel rollouts.

\section{Method}
\label{sec:method}

\begin{figure}[t]
  \centering
  \includegraphics[width=\linewidth]{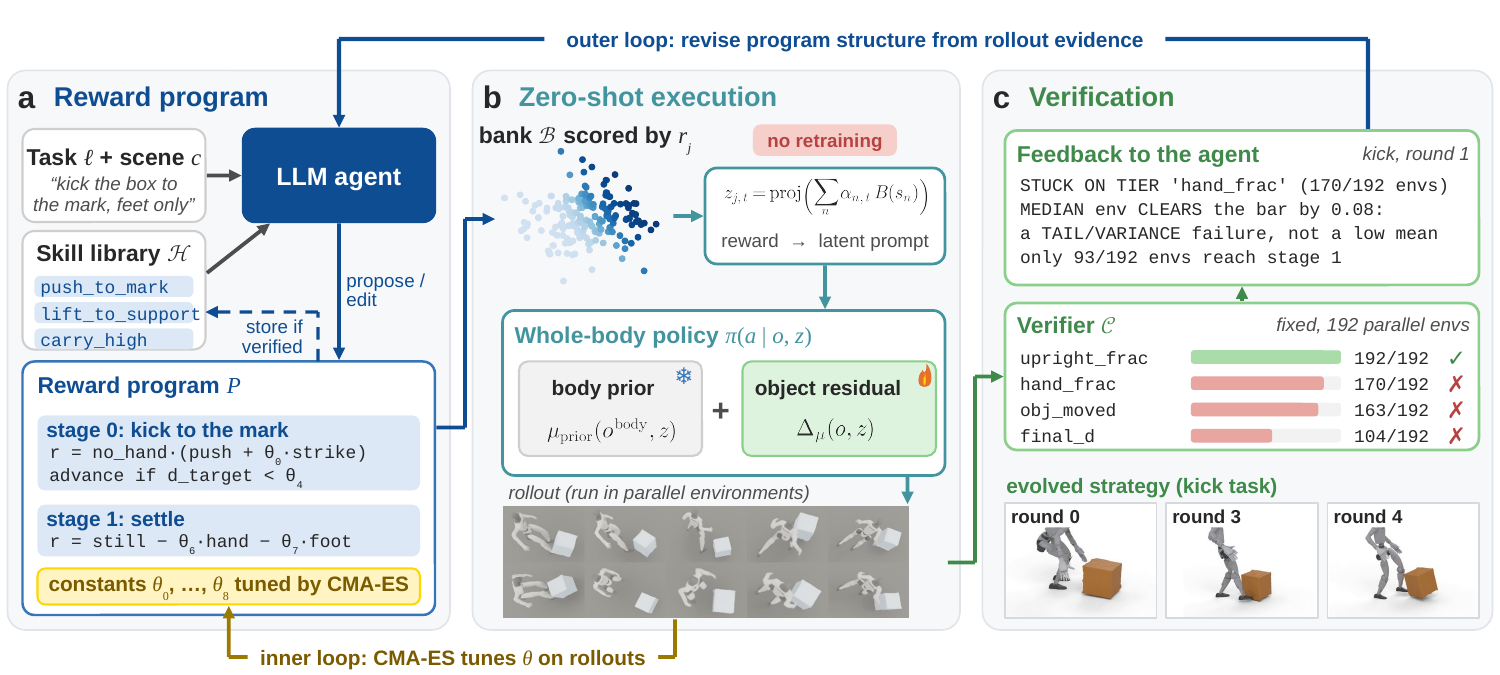}
  \caption{\textbf{InterEvolve overview.} (a) Given a task and scene, an LLM agent writes a staged reward program, drawing on verified programs in the skill library. (b) The forward-backward model scores the active stage's reward over a fixed state bank and projects it into a latent prompt, which a frozen body prior and a pretrained object residual execute without retraining. (c) A fixed verifier scores parallel rollouts, and criterion-level feedback drives the next structural revision (outer loop), while CMA-ES tunes the program constants (inner loop).}
  \label{fig:method}
  \vspace{-1.6em}
\end{figure}

InterEvolve adapts a fixed whole-body controller to new loco-manipulation tasks by evolving a reward programs at test time, and keeps verified programs in a skill library for later tasks to build on (Fig.~\ref{fig:method}). Sec.~\ref{sec:method:setup} formalizes this evolving problem and its budget. Sec.~\ref{sec:method:interface} defines reward programs, the editable task strategies that the evolve operates on. Sec.~\ref{sec:method:fb} shows how an object-aware forward-backward (FB) behavioral foundation model executes a reward program without retraining. Sec.~\ref{sec:method:agent} presents the evolution: an LLM agent revises program structure, CMA-ES~\citep{hansen2001cmaes} calibrates program constants, and the skill library carries verified programs to later tasks.

\subsection{Problem setup}
\label{sec:method:setup}

A task is a language request $\ell$ and a scene context $c$, such as object poses and obstacles. In Figure~\ref{fig:method}, $\ell$ asks the robot to kick a box to a mark with its feet only. A fixed pretrained controller $\pi(\boldsymbol{a}_t\mid\boldsymbol{o}_t,\boldsymbol{z})$ maps policy observation $\boldsymbol{o}_t$ and latent prompt $\boldsymbol{z}$ to action $\boldsymbol{a}_t$. The agent links them through a reward program $P$ of staged rewards (Sec.~\ref{sec:method:interface}), whose active-stage reward becomes $\boldsymbol{z}$ (Sec.~\ref{sec:method:fb}). Executing $P$ from a scenario, a randomized initial condition, yields a trajectory $\tau=(\boldsymbol{s}_0,\boldsymbol{a}_0,\ldots,\boldsymbol{s}_T)$ over the privileged body-object state $\boldsymbol{s}_t$ that rewards and the verifier read.
Success is judged by a verifier $\mathcal C=\{C_i\}_{i=1}^{N_C}$ whose criteria $C_i(\tau)$ check task outcomes and physical constraints, such as staying upright and keeping the hands off the box (Figure~\ref{fig:method}c). For language-specified tasks, an LLM-assisted parser instantiates $\mathcal C$ from $\ell$ and $c$ once, and $\mathcal C$ then stays fixed, so the agent cannot ease a task by rewriting its evaluator. A request thus goes through $(\ell,c)\to P\to\boldsymbol{z}\to\tau\to\mathcal C(\tau)$, and we cast adaptation as a evolution over $P$ alone: within a budget of $K$ rounds, find a program whose trajectories satisfy better criterion in $\mathcal C$. The evolution keeps a \textit{current best program}, the strongest one $\mathcal C$ has confirmed so far, and each round tries to replace it (Sec.~\ref{sec:method:agent}).
Across tasks, a skill library $\mathcal H$ stores each task's final program and verifier results, so the agent can start a new task from earlier solutions, such as the push, lift, and carry programs in Figure~\ref{fig:method}a.

\subsection{Reward programs as editable task strategies}
\label{sec:method:interface}

The evolution needs a strategy representation that can specify multi-stage, contact-rich interaction and can be edited in response to execution evidence. A reward program $P=\big(\boldsymbol{\theta},\{(r_j,g_j)\}_{j=1}^{J}\big)$ consists of an ordered list of $J$ stages and a vector of tunable constants $\boldsymbol{\theta}$. Each stage $j$ is one phase of interaction, such as orienting toward an object, acquiring contact, transporting, or releasing. As shown in Figure~\ref{fig:method}, it holds reward code $r_j$, which scores body-object states through declared features, and a completion condition $g_j$, which reads live rollout context such as object pose, contact status, and stage time and returns whether the stage is complete. Execution stays in stage $j$ until $g_j$ holds and then advances to stage $j+1$, and the last stage runs until the episode ends. The constants $\boldsymbol{\theta}$ are the numbers that $r_j$ and $g_j$ read, such as weights, tolerances, kernel widths, and stage thresholds.

This representation gives the agent explicit handles for evolution: what each stage rewards ($r_j$), when execution moves to the next stage ($g_j$), and the weights and thresholds that calibrate both ($\boldsymbol{\theta}$). The controller, in turn, realizes each stage with the motor behaviors learned in pretraining. To switch from lifting an object to sliding it, for example, the agent edits the object-motion rewards and stage conditions instead of specifying a new joint trajectory. The representation also separates the \textit{structure} of a program, namely its stages, reward terms, and completion conditions, from its constants $\boldsymbol{\theta}$, so the agent revises the structure while a numerical optimizer tunes $\boldsymbol{\theta}$ (Sec.~\ref{sec:method:agent}).

\subsection{Executing reward programs without retraining}
\label{sec:method:fb}
\label{sec:method:arch}

Evaluating each candidate program by training a policy for it would make evolution prohibitively slow. We instead execute programs with an FB behavioral foundation model~\citep{touati2021fb,tirinzoni2025motivo,li2026bfmzero}, which we call the motor model and whose actor is the controller $\pi$ of Sec.~\ref{sec:method:setup}. A latent $\boldsymbol{z}$ indexes a policy, and forward and backward maps factorize the discounted future-state occupancy of that policy
  $M^{\pi_{\boldsymbol{z}}}(\mathrm{d}\boldsymbol{s}'\mid \boldsymbol{s},\boldsymbol{a})
  \approx F(\boldsymbol{s},\boldsymbol{a},\boldsymbol{z})^{\top}B(\boldsymbol{s}')\,\rho(\mathrm{d}\boldsymbol{s}').
  \label{eq:fb}$
For any reward $r$, integrating it against this occupancy gives $Q^{\pi_{\boldsymbol{z}}}_{r}(\boldsymbol{s},\boldsymbol{a})\approx F(\boldsymbol{s},\boldsymbol{a},\boldsymbol{z})^{\top}\widetilde{\boldsymbol{z}}_r$ with $\widetilde{\boldsymbol{z}}_r=\mathbb E_{\rho}[r(\boldsymbol{s}')B(\boldsymbol{s}')]$. A new reward therefore requires only a new prompt for the same policy.

\noindent\textbf{From stage reward to latent prompt.}
We estimate this prompt on a reward-inference bank $\mathcal B=\{\boldsymbol{s}_n\}_{n=1}^{N}$ of body-object states. The bank is sampled once, after motor-model training and before any evolution, from its multi-task, multi-object training replay, which already covers phases such as contact acquisition, transport, and release. For each bank state we cache $B(\boldsymbol{s}_n)$ and the features reward code reads; the bank stays fixed during evolution (Sec.~\ref{sec:app:aggregation}). Let $\hat r_{j,n,t}$ be the reward of stage $j$ on bank state $\boldsymbol{s}_n$, evaluated with the live rollout context at time $t$ (Fig.~\ref{fig:method}b). The prompt is
\begin{equation}
  \boldsymbol{z}_{j,t}=\operatorname{proj}\!\left(
  \sum_{n=1}^{N}\alpha_{j,n,t}\,B(\boldsymbol{s}_n)\right),\qquad
  \alpha_{j,n,t}\propto \hat r_{j,n,t}\exp(\beta\,\hat r_{j,n,t}),
  \label{eq:zr}
\end{equation}
where the normalized weights $\alpha_{j,n,t}$ tilt the estimate toward high-reward states, e.g., a foot striking the box in stage~0 of the kick program. The rest of the bank keeps broad coverage, so other learned behaviors can still support the task (Sec.~\ref{sec:app:inference}). As stage rewards read live context, the prompt is recomputed at every control step and drives the controller $\pi(\boldsymbol{a}_t\mid\boldsymbol{o}_t,\boldsymbol{z}_{j,t})$.

\noindent\textbf{Object-aware interaction representation.}
The pretrained FB model of BFM-Zero~\citep{li2026bfmzero} observes only the body. Rewards that differ only in where the object should go can therefore collapse into the same behavior. We give all three of its networks, the actor and the forward and backward maps, access to the object, while each keeps its pretrained body branch frozen (Fig.~\ref{fig:method}b). The actor keeps the local body observations of BFM-Zero and adds heading-frame object features: object position, orientation, linear and angular velocity, and distance-decayed vectors from body links to the nearest object surface~\citep{xu2025intermimic}. Its mean action adds a trainable object-conditioned residual to the frozen body prior:
$\boldsymbol{\mu}_\pi(\boldsymbol{o}_t,\boldsymbol{z}_{j,t})=\tanh\!\big(\boldsymbol{\mu}_{\mathrm{prior}}(\boldsymbol{o}^{\mathrm{body}}_t,\boldsymbol{z}_{j,t})
  +\boldsymbol{\Delta}_\mu(\boldsymbol{o}_t,\boldsymbol{z}_{j,t})\big).
  \label{eq:residual}$
where $\boldsymbol{o}^{\mathrm{body}}_t$ is the body-only part of $\boldsymbol{o}_t$. The forward and backward maps are extended in the same way, each adding a trainable object residual to a frozen body branch. We train the residual branches on human-object interaction data with the FB objective and a demonstration discriminator (Sec.~\ref{sec:app:motor}), so the motor model becomes object-aware while the frozen prior keeps its body behaviors.

\subsection{Evolving reward programs at test time}
\label{sec:method:agent}

InterEvolve evolves over programs with two nested loops (Algorithm~\ref{alg:search}), because program structure and constants call for different searchers. An LLM agent is well suited to choosing a program's reward terms and stages, whereas a numerical optimizer is better at finding the constants that make a given structure work. The outer loop therefore lets the agent revise program structure, and the inner loop tunes $\boldsymbol{\theta}$ with CMA-ES~\citep{hansen2001cmaes}. Both loops evaluate each candidate on a set of scenarios in parallel, so every decision rests on many rollouts.

\noindent\textbf{Outer loop: structural revision.}
Each round starts with one prompt to the agent (Sec.~\ref{sec:app:context}). The prompt contains the task request, scene context, and verifier $\mathcal C$, the current best program with its rollout feedback, and the skill library $\mathcal H$. The agent learns only from this context, as its weights stay fixed. It proposes several new programs. A validator discards programs that break the code rules, such as reading an undeclared feature, before any rollout. Keeping the current best in context encourages targeted repairs when a strategy is close, while allowing new structures when it is not.

\noindent\textbf{Inner loop: numerical calibration.}
The same structure can fail under poor relative weighting, so each valid proposal is calibrated before judging. The inner loop freezes a program's stages and reward terms and evolves only over $\boldsymbol{\theta}$ (\textit{e.g.} strike weight $\theta_0$ and the stage threshold $\theta_4$ in Figure~\ref{fig:method}a), within agent-declared bounds. CMA-ES samples constants, the simulator evaluates each on the evolution scenarios in parallel, and the sampling distribution moves toward better-scoring constants.

\noindent\textbf{Selection and feedback.}
Tuned candidates are compared with the current best program on the same evolution scenarios, so outcome differences reflect the programs rather than their initial conditions. The strongest candidate is then re-evaluated together with the current best, three times each. It replaces the current best only if its improvement under $\mathcal C$ exceeds the run-to-run evaluation noise. An accepted edit is one whose gain carries over to new initial conditions. The next prompt reports these outcomes per criterion and per stage. For round~1 of the kick task (Figure~\ref{fig:method}c), it reports that the hand criterion still fails in a minority of environments although the median environment passes it, and how each program leads to failure. Such reports locate the stage to repair, and learn from success, as well as failures and their reasons behind (Sec.~\ref{sec:app:context}).

\noindent\textbf{Budget and termination.}
The budget $K$ counts revision rounds after the initial program. Each round makes one agent call and spends 192 tuning rollouts on every valid proposal plus 192 rollouts to confirm the top candidate, so its cost is bounded, and simulation dominates it (Sec.~\ref{sec:app:compute}). Evolution stops once the $K$ rounds are spent, or earlier if the current best program satisfies every criterion in $\mathcal C$ on the confirmation scenarios. It returns the current best program, which is then added to $\mathcal H$.

\noindent\textbf{Skill library.}
The skill library is a text document for summarizing tasks, placed in the agent's prompt every round. Each entry records a completed task: its scene, task text, verifier with the pass rate of each criterion, and the selected program with its tuned constants. With these entries in context, the agent can compose or adapt established programs instead of evolving from scratch.

\section{Experiments}
\label{sec:experiments}

This section verifies that our low-level controller already holds much of the competence a new task needs, and that test-time scaling over reward programs makes it accessible, through three questions. (\textbf{i}) \textit{Competence:} can object-aware motor model provide reusable whole-body interaction, and does better reward improve tracking (Sec.~\ref{sec:exp:tracking})? (\textbf{ii}) \textit{Access:} without reference motion, does execution-guided evolution reach behaviors fixed programs miss, and which evolution components produce the gain (Sec.~\ref{sec:exp:skills})? (\textbf{iii}) \textit{Accumulation:} does a skill library solve composite tasks whose contact modes it covers, and do selected programs transfer to new conditions and hardware (Sec.~\ref{sec:exp:library})?

\subsection{Experimental setup}
\label{sec:exp:setup}

\noindent\textbf{Data and embodiment.}
We pretrain the separated FB models on human-object interaction from OMOMO~\citep{omomo} and GRAB~\citep{grab}. OMOMO covers whole-body manipulation of large objects such as boxes and tables, GRAB whole-body grasping of small objects. We retarget both to the Unitree G1~\citep{unitreeg1} with rubber hands and to the G1 with Inspire hands~\citep{inspirehands}, adapting OmniRetarget~\citep{yang2025omniretarget} to the dexterous hands. Following ULTRA~\citep{he2026ultra}, we use the four box-like OMOMO objects: large box, plastic box, small box, and suitcase. For each object we hold out 50 clips for evaluation and train on the remaining 3{,}866; the held-out large-box clips form the tracking benchmark. The Inspire-hand G1 with GRAB shows qualitatively that InterEvolve extends to dexterous whole-body manipulation. DeepSeek-V4-Flash~\citep{xu2026deepseek} writes the reward programs. All controllers run in Isaac Lab~\citep{mittal2025isaac}, the transfer study replays selected programs on the real G1. For general-purpose tasks, we create a dedicated evaluation benchmark (Sec.~\ref{sec:app:tasks}).

\noindent\textbf{Metrics.}
For tracking, $E_h$ and $E_o$ are the mean body-joint and object-surface errors against the reference in cm, and the success rate (\textbf{SR}) is the fraction of clips tracked to the end without a fall, an object deviation above 0.5\,m, or a lost required contact~\citep{xu2025intermimic}. For general-purpose tasks, SR requires every task criterion and physical constraint to hold in the same rollout, while earned tiers (\textbf{Earned}) measure the mean fraction of criteria passed. Standing still already passes about half of the criteria (Table~\ref{tab:search:comparison}), so SR is the more representative primary metric. Every reported metric is averaged over three independent evaluation runs. We report evolving cost as single-GPU wall time (GPU-h) and total LLM tokens (prompt + output, millions) for one complete evolution on one task family. Fixed external criteria score each general-purpose task; details are in Sec.~\ref{sec:app:eval-details}.

\noindent\textbf{Baselines.}
For tracking, we adapt the tracking policy of ULTRA~\citep{he2026ultra}, a generalist trained with motion imitation to follow loco-manipulation references, which we train on the same data as our motor model, and with BFM-Zero~\citep{li2026bfmzero}, a pretrained body-only FB model. For general-purpose tasks, no public method applies directly to our setting under complex scene context, so we compare InterEvolve with several variations on the same controller: a human-written program and the agent's initial program, each executed directly or after CMA-ES calibration.

\begin{table}[t]
\centering
\small
\begin{minipage}[t]{0.41\linewidth}
\caption{\textbf{Reference tracking} on held-out clips. An evolved reward lifts our success above the dedicated tracker, which keeps the lowest errors; bold marks the best reward-promptable controller. Errors in cm, SR in \%. Evolving: with a reward the agent discovered for tracking.}
\label{tab:tracking:refinement}
\end{minipage}\hfill
\begin{minipage}[t]{0.56\linewidth}
\caption{\textbf{Goal-conditioned tasks}, averaged over eight task families. Evolution more than doubles the success of the best calibrated fixed program. Tune: CMA-ES calibration. Evolving: execution-guided revision. SR and earned tiers in \%, tokens in millions (one family). whereas InterEvolve runs five rounds and calibrates several proposals per round, hence its higher GPU-h.}

\label{tab:search:comparison}
\end{minipage}

\begin{minipage}[t]{0.41\linewidth}
\centering
\setlength{\tabcolsep}{2.2pt}
\begin{tabular}[t]{@{}lcccc@{}}
\toprule
Controller & Evolving & $E_h\downarrow$ & $E_o\downarrow$ & SR$\uparrow$ \\
\midrule
ULTRA & \xmark & \textbf{15.68} & \textbf{21.45} & {66} \\
BFM-Zero & \xmark & 28.36 & 62.33 & {8} \\
InterEvolve & \xmark & 27.20 & 30.80 & {60} \\
\rowcolor{gray!10}
InterEvolve & \cmark & 19.80 & 24.92 & {\textbf{72}} \\
\bottomrule
\end{tabular}
\end{minipage}\hfill
\begin{minipage}[t]{0.56\linewidth}
\centering
\setlength{\tabcolsep}{2.2pt}
\begin{tabular}[t]{@{}lcccccc@{}}
\toprule
Program & Tune & Evolving & SR$\uparrow$ & Earned$\uparrow$ & GPU-h$\downarrow$ & Tokens$\downarrow$ \\
\midrule
Inaction & -- & -- & 0.0 & 49.1 & \notapplicable & \notapplicable \\
Human & \xmark & \xmark & 8.0 & 65.2 & \notapplicable & \notapplicable \\
Agent & \xmark & \xmark & 32.2 & 73.3 & \notapplicable & 0.03 \\
Human & \cmark & \xmark & 18.0 & 75.2 & 0.6 & \notapplicable \\
Agent & \cmark & \xmark & 34.6 & 82.5 & 0.6 & 0.03 \\
\rowcolor{gray!10}
InterEvolve & \cmark & \cmark & \textbf{86.5} & \textbf{95.6} & 2.1 & 0.23 \\
\bottomrule
\end{tabular}
\end{minipage}
\vspace{-0.5em}
\end{table}

\subsection{Reusable control for reference tracking}
\label{sec:exp:tracking}

\noindent\textbf{Object-aware FB can match a dedicated tracker.}
Table~\ref{tab:tracking:refinement} tests whether one controller can mimic body motion and object interaction. ULTRA, an RL tracker trained for tracking alone, has the lowest errors. Our motor model, which also serves reward programs, approaches its success rate, whereas the body-only BFM-Zero loses the object on most clips, so object-aware pretraining supplies the interaction competence that reward programs repurpose. Without the link-to-surface vectors or the object state, the controller still follows the body but rarely succeeds (Table~\ref{tab:control:features}), and the same network trained from scratch almost never works, whereas residuals on the frozen body prior improve with width (Table~\ref{tab:control:capacity}), so the prior carries body skills and the residual learns interaction.

\begin{figure}[t]
\centering
\begin{minipage}[b]{0.507\linewidth}\centering
\includegraphics[width=\linewidth]{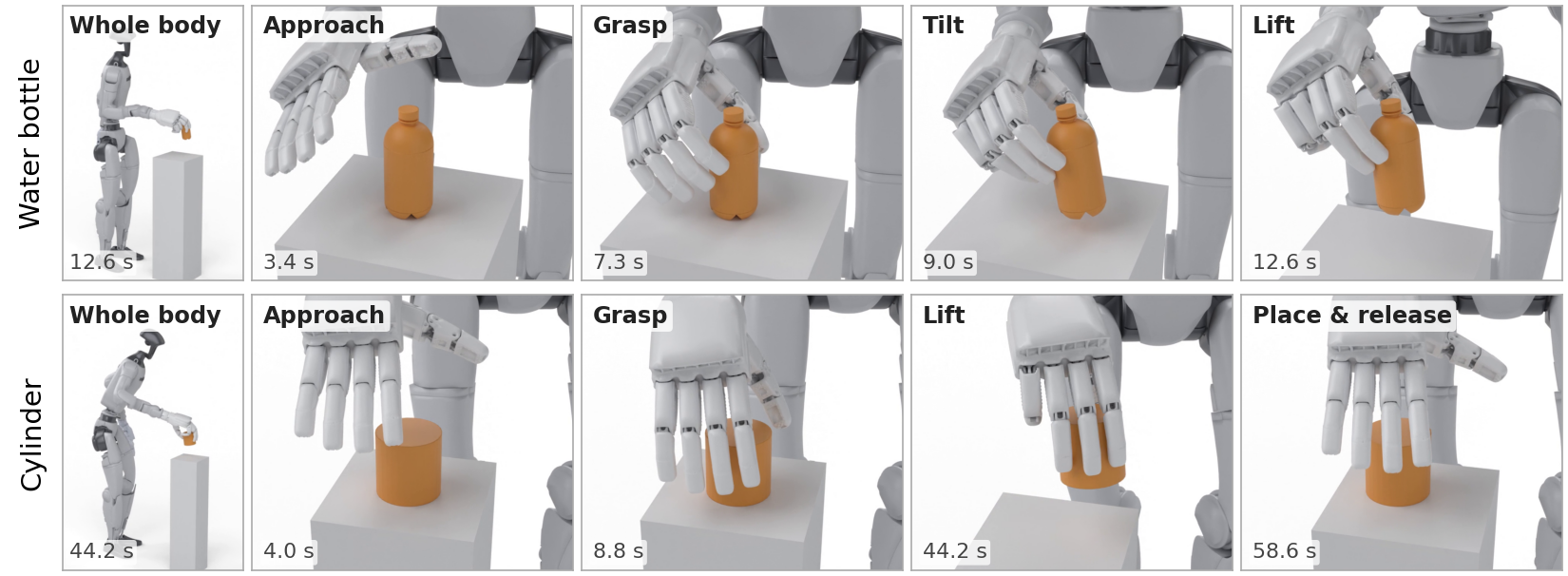}\\[1pt]
{\small (a) Dexterous manipulation of small objects}
\end{minipage}\hfill
\begin{minipage}[b]{0.478\linewidth}\centering
\includegraphics[width=\linewidth]{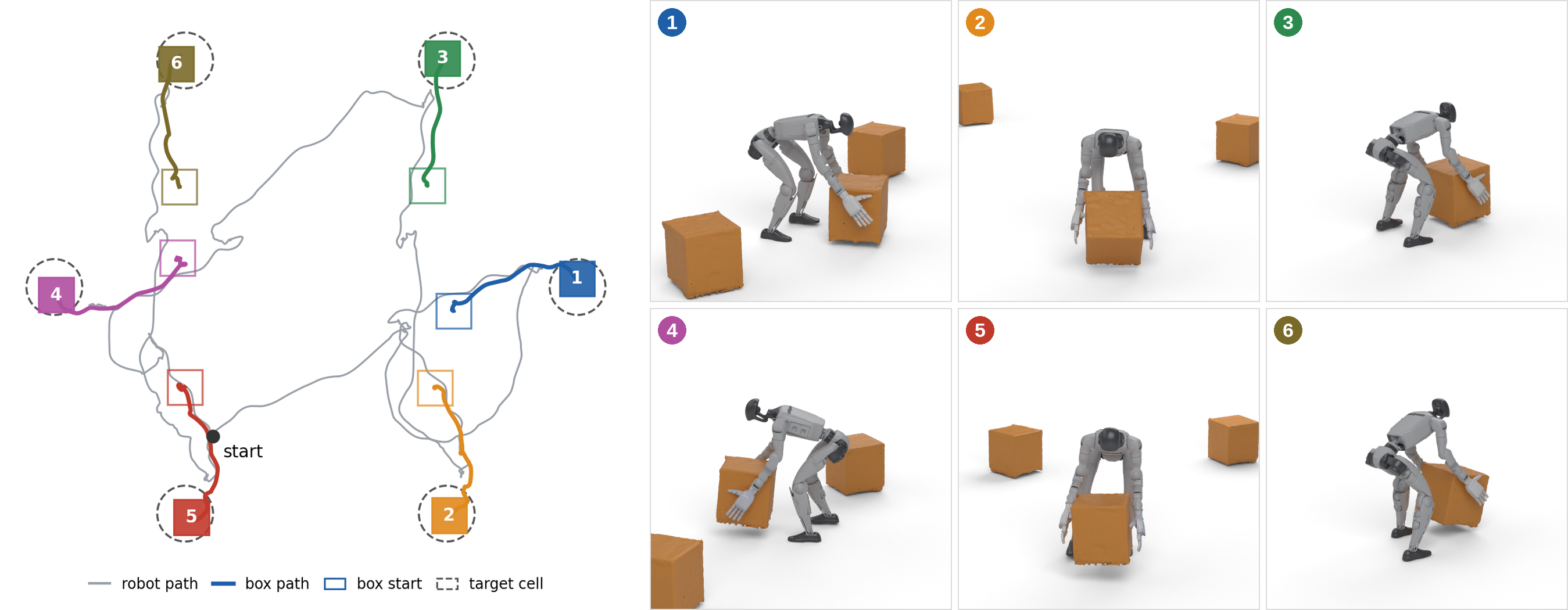}\\[1pt]
{\small (b) Six boxes arranged in one take}
\end{minipage}
\caption{\textbf{Dexterous hands and long-horizon composition.} \textbf{(a)} The G1 with Inspire hands, tracking a reference with the tracking reward calibrated by InterEvolve. \textbf{Left}: whole-body view at the moment of lift; \textbf{Right}:
close-ups of the manipulation phases. \textbf{(b)} Arranging six boxes by the evolution. \textbf{Left}: top-down trace of the robot (gray) and each box from its start (outlined square) to its target cell (dashed circle); numbers give the push order. \textbf{Right}: the robot pushing each box.}
\label{fig:dexcompose}
\vspace{-1.3em}
\end{figure}

\noindent\textbf{Evolving rewards unlock the potential for motion tracking.}
\textit{Reward design matters} even in tracking. Our actor observes the body and object in a local frame, so world-frame drift from the reference is invisible to it. The agent finds a reward that supplies this signal: an object-anchored drift correction blended with the reference prompt (Sec.~\ref{sec:app:drift-correction}). It reduces both errors, lifts success above ULTRA, and nearly halves the final drift of root and object (Figure~\ref{fig:app:drift}). The same procedure also tracks dexterous whole-body manipulation of small objects with Inspire hands (Figure~\ref{fig:dexcompose}(a)).

\subsection{Reward-program evolution for general-purpose tasks}
\label{sec:exp:skills}

We next remove the reference motion and specify only task outcomes and constraints. The benchmark spans eight task families (Sec.~\ref{sec:app:tasks}): pushing to a mark, carrying at several heights, tipping or reorienting an object, feet-only interaction, lifting onto a support, and pushing through gates or around obstacles. Success depends on contact mode, obstacle layout, and when to switch phases. Each run starts from the task text and the observed scene, and only the reward program changes.

\noindent\textbf{Formulation, not hyperparameters, limits a written reward.}
Table~\ref{tab:search:comparison} separates three effects. A reward written once is not enough: even the agent's initial program, which outperforms the human-designed one, fails in most episodes. CMA-ES calibration of the weights alone brings limited gains, since better weights cannot repair an objective with the wrong structure. InterEvolve more than doubles the success of the best calibrated program, and because both fixed programs are calibrated, this gap comes from structural changes: what is rewarded and how objectives are staged.

\noindent\textbf{Evolution helps most where fixed programs fail.}
Calibrated fixed programs solve pushing to a mark but succeed in at most a quarter of the episodes when carrying at chest height, tipping onto a new face, or pushing through a gate or around an obstacle (Table~\ref{tab:app:task-results}). Evolution lifts each of these families above 70\% success. Kicking to a mark remains the hardest, since kicks are rare in the training data and need strong whole-body coordination from the reward.

\begin{table}[t]
\centering
\small
\caption{\textbf{Ablation of the search design.} Stages and numerical calibration are most important designs. Each row disables one component (\xmark) of the full system; definitions are in Sec.~\ref{sec:app:search-ablations}.}

\label{tab:search:mechanisms}
\setlength{\tabcolsep}{3pt}
\begin{tabular*}{0.7\linewidth}{@{\extracolsep{\fill}}ccccccccc@{}}
\toprule
\multicolumn{5}{c}{Search components} & \multicolumn{4}{c}{Results} \\
\cmidrule(r){1-5}\cmidrule(l){6-9}
\shortstack{CMA-ES\\tuning} & \shortstack{Targeted\\edits} & \shortstack{Multi-\\scenario} & \shortstack{Scene\\context} & \shortstack{Multi-\\stage} & \shortstack{SR\\(\%)$\uparrow$} & \shortstack{Earned\\(\%)$\uparrow$} & \shortstack{GPU-h\\$\downarrow$} & \shortstack{Tokens\\(M)$\downarrow$} \\
\midrule
\rowcolor{gray!10}
\cmark & \cmark & \cmark & \cmark & \cmark & \textbf{86.5} & \textbf{95.6} & 2.1 & 0.23 \\
\xmark & \cmark & \cmark & \cmark & \cmark & 51.6 & 89.9 & 1.7 & 0.27 \\
\cmark & \xmark & \cmark & \cmark & \cmark & 68.9 & 92.3 & 2.7 & 0.20 \\
\cmark & \cmark & \xmark & \cmark & \cmark & 68.4 & 89.9 & \textbf{1.0} & \textbf{0.17} \\
\cmark & \cmark & \cmark & \xmark & \cmark & 78.7 & 93.2 & 2.0 & 0.21 \\
\cmark & \cmark & \cmark & \cmark & \xmark & 44.7 & 86.9 & 2.1 & 0.19 \\
\bottomrule
\end{tabular*}
\vspace{-1em}
\end{table}

\noindent\textbf{Staging and calibration matter most.}
Every evolution component contributes (Table~\ref{tab:search:mechanisms}; Sec.~\ref{sec:app:search-ablations} defines each variant). Forcing a single stage causes the largest drop, so contact-rich interaction needs a task decomposed into phases with their own objectives. Removing numerical tuning is nearly as harmful, since a program can only be judged once its constants are calibrated. Dropping targeted edits, so that the agent rewrites each program from scratch, or multi-scenario evaluation, so that each candidate runs from a single start, costs a similar amount: repairing a working program is more reliable than regenerating it, and varied starts tell a robust edit from a lucky one. Scene context, which describes obstacles and supports, has the smallest effect but still helps.

\noindent\textbf{Success grows with test-time scaling, though not monotonically.}
Table~\ref{tab:search:rounds} tracks the current best program over rounds, with the controller weights fixed. The first round brings most of the gain by removing the initial program's coarse failures, and later rounds resolve the remaining criteria. Earned tiers rise monotonically, whereas full success dips from round 2 to round 3, since it requires all criteria in the same rollout and fixing one criterion can briefly break another. Tokens grow linearly with rounds, but simulation dominates the cost, with minutes per round in the language model against tens of minutes of rollouts (Sec.~\ref{sec:app:compute}), so a faster simulator such as mjlab~\citep{zakka2026mjlab} in place of Isaac Lab~\citep{mittal2025isaac} could shorten evolution substantially. Figure~\ref{fig:evolution-trace} shows that revised rewards change the contact strategy itself.

\begin{figure}[t]
  \centering
  \includegraphics[width=\linewidth]{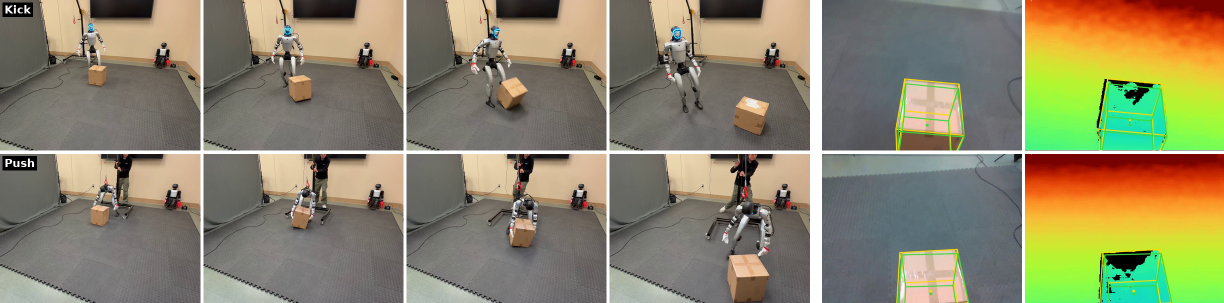}
  \caption{\textbf{Autonomous} execution of real-world G1 with \textbf{onboard camera} and task intent of kicking a box (top) and pushing a box (bottom). Each row runs a program evolved in simulation. The first four columns show the run from a third-person view, left to right in time. The last two columns are the robot's onboard color and depth images as it approaches the box, overlaid with the FoundationPose estimate (yellow) and the filtered pose that the controller receives (green).}
  \label{fig:real}
  \vspace{-1em}
\end{figure}

\noindent\textbf{Qualitative results and real deployment from onboard perception.}
Figures~\ref{fig:case},~\ref{fig:push}, and~\ref{fig:app:variation} show goal-conditioned skills evolved in simulation, in which acquisition, transport, placement, and handling arise from reward programs run by the shared controller. Figure~\ref{fig:real} deploys programs evolved for two tasks on a physical Unitree G1. The robot detects the box with its egocentric camera, and FoundationPose~\citep{wen2024foundationpose} estimates its 6-D pose, which supplies the controller's object features. Each program is evolved and verified in simulation, then run on the robot.

\begin{table}[t]
\centering
\small
\begin{minipage}[t]{0.40\linewidth}
\caption{\textbf{Ablation on search rounds.} Most of the gain comes in round~1; earned tiers rise monotonically. The final row repeats Table~\ref{tab:search:comparison}.}
\label{tab:search:rounds}
\centering
\setlength{\tabcolsep}{4pt}
\begin{tabular}[t]{@{}lccc@{}}
\toprule
Round & SR$\uparrow$ & Earned$\uparrow$ & Tokens \\
\midrule
Initial & 32.2 & 73.3 & 0.03 \\
1 & 75.0 & 91.7 & 0.08 \\
2 & 76.8 & 92.6 & 0.13 \\
3 & 73.6 & 93.1 & 0.18 \\
\rowcolor{gray!10}
4 (final) & \textbf{86.5} & \textbf{95.6} & 0.23 \\
\bottomrule
\end{tabular}
\end{minipage}\hfill
\begin{minipage}[t]{0.56\linewidth}
\caption{\textbf{Reuse on composite tasks} unseen by every library, under one target-task budget. The full library solves the most episodes. Carry-only keeps the three carrying programs. Tokens (millions, mean over the tasks) exclude library construction. Every condition searches for three rounds.}
\label{tab:search:library}
\centering
\setlength{\tabcolsep}{4pt}
\begin{tabular}[t]{@{}lcccc@{}}
\toprule
Library & Relocate & Stack & Carry-place-kick & Tokens$\downarrow$ \\
\midrule
\rowcolor{gray!10}
Full & \textbf{8/10} & \textbf{4/10} & \textbf{5/10} & 0.16 \\
None & 0/10 & 0/10 & 1/10 & 0.13 \\
Carry-only & 6/10 & 1/10 & 1/10 & 0.15 \\
\bottomrule
\end{tabular}
\end{minipage}
\vspace{-1em}
\end{table}

\subsection{Reuse and composition}
\label{sec:exp:library}

\noindent\textbf{Composite tasks need a library that covers their contact modes.}
Evolved experience is valuable only if later tasks can reuse it. Table~\ref{tab:search:library} evaluates three composite tasks that each combine previously evolved objectives in one episode: relocating an object between supports, stacking one box on another, and carrying, placing, then kicking an object. We compare the full skill library with no library and a carry-only library. With the full library, the agent solves most relocation episodes and about half of the stacking and carry-place-kick episodes, but almost none without a library. The carry-only library recovers most of relocation but helps little on stacking or kicking. Reuse therefore depends on covering the contact modes a task needs, not on having any stored experience.

\noindent\textbf{Evolving adapts stored programs.}
A finer ablation on carry, place, and kick shows that stored programs run without evolution almost never solve the task (Table~\ref{tab:app:program-library}), so the gain comes from evolution integrating skill library rather than from direct retrieval. Broader coverage still helps.

\noindent\textbf{Evolution for long-horizon tasks.}
Figure~\ref{fig:dexcompose}(b) chains searched programs to arrange six scattered boxes into a ring. The same agent for evolution orders the boxes, and each leg runs a walking program to the box followed by the evolved pushing program. One controller executes all twelve phases from a single reset, so errors accumulate across legs instead of being reset away. In the take shown, all six boxes end within 11\,cm of their cells and no placed box is disturbed by later legs. More are on \href{https://sirui-xu.github.io/InterEvolve}{project page}.

\section{Discussion and Conclusion}
\label{sec:discussion}

InterEvolve suggests a different division of labor for humanoid intelligence. A behavioral foundation model learns how to move once, and task knowledge lives outside its weights, in reward programs that a language model can read, edit, and test by execution. Reasoning and control thus improve on separate timescales: the controller with more interaction data, the agent with more test-time search and a growing program library, and neither requires retraining the other. Physical execution grounds the agent's reasoning, since it revises what it asks of the controller from observed outcomes rather than assumptions. We see this as a step toward humanoids that, like language models at inference time, improve by testing more, and accumulate skills as inspectable, reusable programs.

\clearpage
{\small\noindent\textbf{Acknowledgments.} We thank Derek Zhang and Lusen Zhao for helping with the on-site deployment.\par}

\bibliography{refs}

@article{andrychowicz2020learning,
  title={Learning dexterous in-hand manipulation},
  author={Andrychowicz, OpenAI: Marcin and Baker, Bowen and Chociej, Maciek and Jozefowicz, Rafal and McGrew, Bob and Pachocki, Jakub and Petron, Arthur and Plappert, Matthias and Powell, Glenn and Ray, Alex and others},
  journal={The International Journal of Robotics Research},
  volume={39},
  number={1},
  pages={3--20},
  year={2020},
  publisher={SAGE Publications Sage UK: London, England}
}

@article{zhang2026sumo,
  title   = {{Sumo}: Dynamic and Generalizable Whole-Body Loco-Manipulation},
  author  = {Zhang, John Z. and Sorokin, Maks and Br{\"u}digam, Jan and Hung, Brandon and Phillips, Stephen and Yershov, Dmitry and Niroui, Farzad and Zhao, Tong and Fermoselle, Leonor and Zhu, Xinghao and Cao, Chao and Ta, Duy and Pang, Tao and Wang, Jiuguang and Culbertson, Preston and Manchester, Zachary and Le Cl{\'e}ac'h, Simon},
  journal = {arXiv preprint arXiv:2604.08508},
  year    = {2026}
}

@article{taouil2026motiondisco,
  title   = {{MotionDisco}: Motion Discovery for Extreme Humanoid Loco-Manipulation},
  author  = {Taouil, Ilyass and Ciebelski, Michal and Omar, Shafeef and Zhao, Haizhou and Dai, Angela and Johnson, Aaron M. and Khadiv, Majid},
  journal = {arXiv preprint arXiv:2606.06139},
  year    = {2026}
}

@article{lee2026rda,
  title   = {{RDA}: Reward Design Agent for Reinforcement Learning},
  author  = {Lee, Hojoon and Subramanian, Ajay and Abbatematteo, Ben and Veerabadran, Vijay and Matias, Pedro and Ridgeway, Karl and Kamra, Nitin},
  journal = {arXiv preprint arXiv:2606.01672},
  year    = {2026}
}

@article{sun2026choreo,
  title   = {{CHOREO}: Every Humanoid Skill as a Trajectory},
  author  = {Sun, Ziyi and Chen, Jingwen and Wang, Yuxi and Xia, Xiuze and Cheng, Long and Zhang, Zhaoxiang and Dong, Yujun},
  journal = {arXiv preprint arXiv:2609.22274},
  year    = {2026}
}

@article{ren2026cybowaiter,
  title   = {{Cybo-Waiter}: A Physical Agentic Framework for Humanoid Whole-Body Locomotion-Manipulation},
  author  = {Ren, Peng and Ge, Haoyang and Qi, Chuan and Huang, Cong and Li, Hong and Zhao, Jiang and Chi, Pei and Chen, Kai},
  journal = {arXiv preprint arXiv:2603.10675},
  year    = {2026}
}

@article{yuan2025being0,
  title   = {{Being-0}: A Humanoid Robotic Agent with Vision-Language Models and Modular Skills},
  author  = {Yuan, Haoqi and Bai, Yu and Fu, Yuhui and Zhou, Bohan and Feng, Yicheng and Xu, Xinrun and Zhan, Yi and Karlsson, B{\"o}rje F. and Lu, Zongqing},
  journal = {arXiv preprint arXiv:2503.12533},
  year    = {2025}
}

@article{wen2025coa,
  title   = {Humanoid Agent via Embodied Chain-of-Action Reasoning with Multimodal Foundation Models for Zero-Shot Loco-Manipulation},
  author  = {Wen, Congcong and Bethala, Geeta Chandra Raju and Hao, Yu and Pudasaini, Niraj and Huang, Hao and Yuan, Shuaihang and Huang, Baoru and Nguyen, Anh and Wang, Mengyu and Tzes, Anthony and Fang, Yi},
  journal = {arXiv preprint arXiv:2504.09532},
  year    = {2025}
}

@article{rupf2025opti,
  title   = {Optimistic Task Inference for Behavior Foundation Models},
  author  = {Rupf, Thomas and Bagatella, Marco and Vlastelica, Marin and Krause, Andreas},
  journal = {arXiv preprint arXiv:2510.20264},
  year    = {2025}
}

@article{bagot2026exploration,
  title   = {Exploration and Online Transfer with Behavioral Foundation Models},
  author  = {Bagot, Louis and Lefort, Mathieu and Matignon, La{\"e}titia},
  journal = {arXiv preprint arXiv:2606.29980},
  year    = {2026}
}

@inproceedings{sikchi2025rlzero,
  title     = {{RLZero}: Direct Policy Inference from Language Without In-Domain Supervision},
  author    = {Sikchi, Harshit and Agarwal, Siddhant and Jajoo, Pranaya and Parajuli, Samyak and Chuck, Caleb and Rudolph, Max and Stone, Peter and Zhang, Amy and Niekum, Scott},
  booktitle = {NeurIPS},
  year      = {2025}
}

@article{cao2026texedo,
  title   = {{TEXEDO}: Test Time Scaling for Controller-aware Language-conditioned Humanoid Motion Generation},
  author  = {Cao, Jianuo and Chen, Yuxin and Song, Yuzhen and Tomizuka, Masayoshi and Li, Chenran and Tian, Thomas},
  journal = {arXiv preprint arXiv:2606.22998},
  year    = {2026}
}

@article{seo2026rgb,
  title   = {{RGB}: {RL} Guided Whole-Body {MPPI} for Humanoid Control},
  author  = {Seo, Yunsoo and Choi, Sol and Im, Euncheol and Lim, Myo Taeg and Lee, Yisoo},
  journal = {arXiv preprint arXiv:2606.25123},
  year    = {2026}
}

@inproceedings{xiao2023unihsi,
  title={Unified human-scene interaction via prompted chain-of-contacts},
  author={Xiao, Zeqi and Wang, Tai and Wang, Jingbo and Cao, Jinkun and Zhang, Wenwei and Dai, Bo and Lin, Dahua and Pang, Jiangmiao},
  booktitle={ICLR},
  year={2024}
}

@inproceedings{tevet2024closd,
  title={{CLoSD}: Closing the loop between simulation and diffusion for multi-task character control},
  author={Tevet, Guy and Raab, Sigal and Cohan, Setareh and Reda, Daniele and Luo, Zhengyi and Peng, Xue Bin and Bermano, Amit and Van de Panne, Michiel},
  booktitle={ICLR},
  year={2025}
}

@article{wu2025stride,
  title   = {{STRIDE}: Automating Reward Design, Deep Reinforcement Learning Training and Feedback Optimization in Humanoid Robotics Locomotion},
  author  = {Wu, Zhenwei and Lu, Jinxiong and Chen, Yuxiao and Liu, Yunxin and Zhuang, Yueting and Hu, Luhui},
  journal = {arXiv preprint arXiv:2502.04692},
  year    = {2025}
}

@article{liang2024lmpc,
  title={Learning to learn faster from human feedback with language model predictive control},
  author={Liang, Jacky and Xia, Fei and Yu, Wenhao and Zeng, Andy and Arenas, Montserrat Gonzalez and Attarian, Maria and Bauza, Maria and Bennice, Matthew and Bewley, Alex and Dostmohamed, Adil and others},
  journal={arXiv preprint arXiv:2402.11450},
  year={2024}
}

@article{wang2026shaper,
  title   = {Self-Evolving Embodied Agents via Skill-Harness Evolution},
  author  = {Wang, Peidong and Ma, Zhiming and Chang, Ying and Luo, Xufang and Zhang, Yiqun and Wang, Zihan and Yang, Xiaocui and Feng, Shi and Yang, Yuqing and Li, Dongsheng},
  journal = {arXiv preprint arXiv:2608.11350},
  year    = {2026}
}

@inproceedings{wen2024foundationpose,
  title={Foundationpose: Unified 6d pose estimation and tracking of novel objects},
  author={Wen, Bowen and Yang, Wei and Kautz, Jan and Birchfield, Stan},
  booktitle={CVPR},
  year={2024},
}

@article{zakka2026mjlab,
  title={mjlab: A lightweight framework for gpu-accelerated robot learning},
  author={Zakka, Kevin and Liao, Qiayuan and Yi, Brent and Lay, Louis Le and Sreenath, Koushil and Abbeel, Pieter},
  journal={arXiv preprint arXiv:2601.22074},
  year={2026}
}

@article{xu2026deepseek,
  title={Deepseek-v4: Towards highly efficient million-token context intelligence},
  author={Xu, Anyi and Lin, Bangcai and Xue, Bing and Wang, Bingxuan and Xu, Bingzheng and Wu, Bochao and Zhang, Bowei and Lin, Chaofan and Dong, Chen and Ling, Chenchen and others},
  journal={arXiv preprint arXiv:2606.19348},
  year={2026}
}

@article{guo2025deepseekr1,
  title={Deepseek-r1: Incentivizing reasoning capability in llms via reinforcement learning},
  author={Guo, Daya and Yang, Dejian and Zhang, Haowei and Song, Junxiao and Wang, Peiyi and Zhu, Qihao and Xu, Runxin and Zhang, Ruoyu and Ma, Shirong and Bi, Xiao and others},
  journal={arXiv preprint arXiv:2501.12948},
  year={2025}
}

@article{novikov2025alphaevolve,
  title={{AlphaEvolve}: A coding agent for scientific and algorithmic discovery},
  author={Novikov, Alexander and V{\~u}, Ng{\^a}n and Eisenberger, Marvin and Dupont, Emilien and Huang, Po-Sen and Wagner, Adam Zsolt and Shirobokov, Sergey and Kozlovskii, Borislav and Ruiz, Francisco JR and Mehrabian, Abbas and others},
  journal={arXiv preprint arXiv:2506.13131},
  year={2025}
}

@inproceedings{hansen2024tdmpc2,
  title={Td-mpc2: Scalable, robust world models for continuous control},
  author={Hansen, Nick and Su, Hao and Wang, Xiaolong},
  booktitle={ICLR},
  year={2024}
}

@inproceedings{
snell2025scaling,
title={Scaling {LLM} Test-Time Compute Optimally Can be More Effective than Scaling Parameters for Reasoning},
author={Charlie Victor Snell and Jaehoon Lee and Kelvin Xu and Aviral Kumar},
booktitle={ICLR},
year={2025},
}

@article{brown2024large,
  title={Large language monkeys: Scaling inference compute with repeated sampling},
  author={Brown, Bradley and Juravsky, Jordan and Ehrlich, Ryan and Clark, Ronald and Le, Quoc V and R{\'e}, Christopher and Mirhoseini, Azalia},
  journal={arXiv preprint arXiv:2407.21787},
  year={2024}
}

@article{omomo,
  title = {Object Motion Guided Human Motion Synthesis},
  author = {Li, Jiaman and Wu, Jiajun and Liu, C. Karen},
  journal = {ACM Transactions on Graphics},
  year = {2023}
}

@inproceedings{grab,
  title = {{GRAB}: A Dataset of Whole-Body Human Grasping of Objects},
  author = {Taheri, Omid and Ghorbani, Nima and Black, Michael J. and Tzionas, Dimitrios},
  booktitle = {ECCV},
  year = {2020}
}

@article{mittal2025isaac,
  title={Isaac lab: A gpu-accelerated simulation framework for multi-modal robot learning},
  author={Mittal, Mayank and Roth, Pascal and Tigue, James and Richard, Antoine and Zhang, Octi and Du, Peter and Serrano-Munoz, Antonio and Yao, Xinjie and Zurbr{\"u}gg, Ren{\'e} and Rudin, Nikita and others},
  journal={arXiv preprint arXiv:2511.04831},
  year={2025}
}

@misc{unitreeg1,
  author = {{Unitree Robotics}},
  title = {{Unitree G1}},
  url = {https://www.unitree.com/g1/},
}

@misc{inspirehands,
  author = {{Inspire Robots}},
  title = {{The Dexterous Hands}},
  url = {https://en.inspire-robots.com/product-category/the-dexterous-hands/},
}

@article{wang2026humanx,
  title={{HumanX}: Toward Agile and Generalizable Humanoid Interaction Skills from Human Videos},
  author={Wang, Yinhuai and Zhao, Qihan and Lau, Yuen Fui and Yu, Runyi and Tsui, Hok Wai and Chen, Qifeng and Wang, Jingbo and Pang, Jiangmiao and Tan, Ping},
  journal={arXiv preprint arXiv:2602.02473},
  year={2026}
}

@article{dayan1993successor,
  title   = {Improving Generalization for Temporal Difference Learning: The Successor Representation},
  author  = {Dayan, Peter},
  journal = {Neural Computation},
  volume  = {5},
  number  = {4},
  year    = {1993}
}

@inproceedings{barreto2017successor,
  title     = {Successor Features for Transfer in Reinforcement Learning},
  author={Barreto, Andr{\'e} and Dabney, Will and Munos, R{\'e}mi and Hunt, Jonathan J and Schaul, Tom and Van Hasselt, Hado and Silver, David},
  booktitle = {NeurIPS},
  year      = {2017}
}

@inproceedings{touati2021fb,
  title     = {Learning One Representation to Optimize All Rewards},
  author    = {Touati, Ahmed and Ollivier, Yann},
  booktitle = {NeurIPS},
  year      = {2021}
}

@inproceedings{touati2023zeroshot,
  title     = {Does Zero-Shot Reinforcement Learning Exist?},
  author    = {Touati, Ahmed and Rapin, J\'er\'emy and Ollivier, Yann},
  booktitle = {ICLR},
  year      = {2023}
}

@article{cetin2025finer,
  title   = {Finer Behavioral Foundation Models via Auto-Regressive Features and Advantage Weighting},
  author  = {Cetin, Edoardo and Touati, Ahmed and Ollivier, Yann},
  journal = {Reinforcement Learning Journal},
  volume  = {6},
  year    = {2025}
}

@inproceedings{bagatella2026tdjepa,
  title     = {{TD-JEPA}: Latent-Predictive Representations for Zero-Shot Reinforcement Learning},
  author    = {Bagatella, Marco and Pirotta, Matteo and Touati, Ahmed and Lazaric, Alessandro and Tirinzoni, Andrea},
  booktitle = {ICLR},
  year      = {2026}
}

@article{peng2022ase,
  title   = {{ASE}: Large-Scale Reusable Adversarial Skill Embeddings for Physically Simulated Characters},
  author  = {Peng, Xue Bin and Guo, Yunrong and Halper, Lina and Levine, Sergey and Fidler, Sanja},
  journal = {ACM Transactions on Graphics},
  year    = {2022}
}

@inproceedings{fu2024humanplus,
  title     = {{HumanPlus}: Humanoid Shadowing and Imitation from Humans},
  author    = {Fu, Zipeng and Zhao, Qingqing and Wu, Qi and Wetzstein, Gordon and Finn, Chelsea},
  booktitle = {CoRL},
  year      = {2024}
}

@inproceedings{he2024omnih2o,
  title     = {{OmniH2O}: Universal and Dexterous Human-to-Humanoid Whole-Body Teleoperation and Learning},
  author    = {He, Tairan and Luo, Zhengyi and He, Xialin and Xiao, Wenli and Zhang, Chong and Zhang, Weinan and Kitani, Kris M. and Liu, Changliu and Shi, Guanya},
  booktitle = {CoRL},
  year      = {2024}
}

@article{chen2025gmt,
  title   = {{GMT}: General Motion Tracking for Humanoid Whole-Body Control},
  author  = {Chen, Zixuan and Ji, Mazeyu and Cheng, Xuxin and Peng, Xuanbin and Peng, Xue Bin and Wang, Xiaolong},
  journal = {arXiv preprint arXiv:2506.14770},
  year    = {2025}
}

@article{yin2026unitracker,
  title   = {{UniTracker}: Learning Universal Whole-Body Motion Tracker for Humanoid Robots},
  author  = {Yin, Kangning and Zeng, Weishuai and Fan, Ke and Dai, Minyue and Wang, Zirui and Zhang, Qiang and Tian, Zheng and Wang, Jingbo and Pang, Jiangmiao and Zhang, Weinan},
  journal = {IEEE Robotics and Automation Letters},
  year    = {2026}
}

@article{liao2025beyondmimic,
  title={Beyondmimic: From motion tracking to versatile humanoid control via guided diffusion},
  author={Liao, Qiayuan and Truong, Takara E and Huang, Xiaoyu and Gao, Yuman and Tevet, Guy and Sreenath, Koushil and Liu, C Karen},
  journal={Science Robotics},
  volume={11},
  number={117},
  pages={eadx8924},
  year={2026},
  publisher={American Association for the Advancement of Science}
}

@article{sonic2025,
  title   = {{SONIC}: Supersizing Motion Tracking for Natural Humanoid Whole-Body Control},
  author  = {Luo, Zhengyi and Yuan, Ye and Wang, Tingwu and Li, Chenran and Casta{\~n}eda, Fernando and Chen, Sirui and Cao, Zi-Ang and Li, Jiefeng and Minor, David and Ben, Qingwei and Park, Jinhyung and Sami, David and Wang, Zi and Da, Xingye and Ding, Runyu and Hogg, Cyrus and Song, Lina and Lim, Edy and Jeong, Eugene and He, Tairan and Xue, Haoru and Xiao, Wenli and Yuen, Simon and Kautz, Jan and Chang, Yan and Iqbal, Umar and Fan, Linxi Jim and Zhu, Yuke},
  journal = {Science Robotics},
  volume  = {11},
  number  = {117},
  pages   = {eaed4592},
  year    = {2026},
  doi     = {10.1126/scirobotics.aed4592}
}

@inproceedings{ze2025twist,
  title     = {{TWIST}: Teleoperated Whole-Body Imitation System},
  author    = {Ze, Yanjie and Chen, Zixuan and Araujo, Joao Pedro and Cao, Zi-Ang and Peng, Xue Bin and Wu, Jiajun and Liu, C. Karen},
  booktitle = {CoRL},
  year      = {2025}
}

@inproceedings{kareer2024egomimic,
  title   = {{EgoMimic}: Scaling Imitation Learning via Egocentric Video},
  author  = {Kareer, Simar and Patel, Dhruv and Punamiya, Ryan and Mathur, Pranay and Cheng, Shuo and Wang, Chen and Hoffman, Judy and Xu, Danfei},
  booktitle = {ICRA},
  year    = {2025}
}

@article{ji2024exbody2,
  title   = {{ExBody2}: Advanced Expressive Humanoid Whole-Body Control},
  author  = {Ji, Mazeyu and Peng, Xuanbin and Liu, Fangchen and Li, Jialong and Yang, Ge and Cheng, Xuxin and Wang, Xiaolong},
  journal = {arXiv preprint arXiv:2412.13196},
  year    = {2024}
}

@inproceedings{li2024okami,
  title     = {{OKAMI}: Teaching Humanoid Robots Manipulation Skills through Single Video Imitation},
  author    = {Li, Jinhan and Zhu, Yifeng and Xie, Yuqi and Jiang, Zhenyu and Seo, Mingyo and Pavlakos, Georgios and Zhu, Yuke},
  booktitle = {CoRL},
  year      = {2024}
}

@article{liu2024opt2skill,
  title   = {{Opt2Skill}: Imitating Dynamically-Feasible Whole-Body Trajectories for Versatile Humanoid Loco-Manipulation},
  author  = {Liu, Fukang and Gu, Zhaoyuan and Cai, Yilin and Zhou, Ziyi and Jung, Hyunyoung and Jang, Jaehwi and Zhao, Shijie and Ha, Sehoon and Chen, Yue and Xu, Danfei and Zhao, Ye},
  journal = {IEEE Robotics and Automation Letters},
  year    = {2025}
}

@inproceedings{luo2024omnigrasp,
  title     = {{OmniGrasp}: Grasping Diverse Objects with Simulated Humanoids},
  author    = {Luo, Zhengyi and Cao, Jinkun and Christen, Sammy and Winkler, Alexander and Kitani, Kris M. and Xu, Weipeng},
  booktitle = {NeurIPS},
  year      = {2024}
}

@inproceedings{tirinzoni2025motivo,
  title     = {Zero-Shot Whole-Body Humanoid Control via Behavioral Foundation Models},
  author    = {Tirinzoni, Andrea and Touati, Ahmed and Farebrother, Jesse and Guzek, Mateusz and Kanervisto, Anssi and Xu, Yingchen and Lazaric, Alessandro and Pirotta, Matteo},
  booktitle = {ICLR},
  year      = {2025}
}

@inproceedings{li2026bfmzero,
  title     = {{BFM-Zero}: A Promptable Behavioral Foundation Model for Humanoid Control Using Unsupervised Reinforcement Learning},
  author    = {Li, Yitang and Luo, Zhengyi and Zhang, Tonghe and Dai, Cunxi and Kanervisto, Anssi and Tirinzoni, Andrea and Weng, Haoyang and Kitani, Kris M. and Guzek, Mateusz and Touati, Ahmed and Lazaric, Alessandro and Pirotta, Matteo and Shi, Guanya},
  booktitle = {ICLR},
  year      = {2026}
}

@article{sikchi2025fastadapt,
  title   = {Fast Adaptation with Behavioral Foundation Models},
  author  = {Sikchi, Harshit and Tirinzoni, Andrea and Touati, Ahmed and Xu, Yingchen and Kanervisto, Anssi and Niekum, Scott and Zhang, Amy and Lazaric, Alessandro and Pirotta, Matteo},
  journal = {arXiv preprint arXiv:2504.07896},
  year    = {2025}
}

@article{chen2026imagine2real,
  title   = {{Imagine2Real}: Towards Zero-Shot Humanoid-Object Interaction via Video Generative Priors},
  author  = {Chen, Jiahe and Wang, ZiRui and Jia, Feiyu and Chen, Xiao and Niu, Xiaojie and Zeng, Weishuai and Xue, Tianfan and Zhou, Xiaowei and Pang, Jiangmiao and Wang, Jingbo},
  journal = {arXiv preprint arXiv:2605.22272},
  year    = {2026}
}

@inproceedings{xu2025intermimic,
  title     = {{InterMimic}: Towards Universal Whole-Body Control for Physics-Based Human-Object Interactions},
  author    = {Xu, Sirui and Ling, Hung Yu and Wang, Yu-Xiong and Gui, Liang-Yan},
  booktitle = {CVPR},
  year      = {2025}
}

@article{xu2026interprior,
  title   = {{InterPrior}: Scaling Generative Control for Physics-Based Human-Object Interactions},
  author  = {Xu, Sirui and Schulter, Samuel and Ziyadi, Morteza and He, Xialin and Fei, Xiaohan and Wang, Yu-Xiong and Gui, Liang-Yan},
  journal = {arXiv preprint arXiv:2602.06035},
  year    = {2026}
}

@inproceedings{wang2025skillmimic,
  title     = {{SkillMimic}: Learning Basketball Interaction Skills from Demonstrations},
  author    = {Wang, Yinhuai and Zhao, Qihan and Yu, Runyi and Tsui, Hok Wai and Zeng, Ailing and Lin, Jing and Luo, Zhengyi and Yu, Jiwen and Li, Xiu and Chen, Qifeng and Zhang, Jian and Zhang, Lei and Tan, Ping},
  booktitle = {CVPR},
  year      = {2025}
}

@article{sun2025ulc,
  title   = {{ULC}: A Unified and Fine-Grained Controller for Humanoid Loco-Manipulation},
  author  = {Sun, Wandong and Feng, Luying and Cao, Baoshi and Liu, Yang and Jin, Yaochu and Xie, Zongwu},
  journal = {arXiv preprint arXiv:2507.06905},
  year    = {2025}
}

@article{yang2025omniretarget,
  title   = {{OmniRetarget}: Interaction-Preserving Data Generation for Humanoid Whole-Body Loco-Manipulation and Scene Interaction},
  author  = {Yang, Lujie and Huang, Xiaoyu and Wu, Zhen and Kanazawa, Angjoo and Abbeel, Pieter and Sferrazza, Carmelo and Liu, C. Karen and Duan, Rocky and Shi, Guanya},
  journal = {arXiv preprint arXiv:2509.26633},
  year    = {2025}
}

@article{fu2025demohlm,
  title   = {{DemoHLM}: From One Demonstration to Generalizable Humanoid Loco-Manipulation},
  author  = {Fu, Yuhui and Xie, Feiyang and Xu, Chaoyi and Xiong, Jing and Yuan, Haoqi and Lu, Zongqing},
  journal = {IEEE Robotics and Automation Letters},
  volume  = {11},
  number  = {4},
  pages   = {4393--4400},
  year    = {2026}
}

@article{wang2025physhsi,
  title   = {{PhysHSI}: Towards a Real-World Generalizable and Natural Humanoid-Scene Interaction System},
  author  = {Wang, Huayi and Zhang, Wentao and Yu, Runyi and Huang, Tao and Ren, Junli and Jia, Feiyu and Wang, Zirui and Niu, Xiaojie and Chen, Xiao and Chen, Jiahe and others},
  journal = {arXiv preprint arXiv:2510.11072},
  year    = {2025}
}

@inproceedings{tessler2025maskedmanipulator,
  title={{Maskedmanipulator}: Versatile whole-body control for loco-manipulation},
  author={Tessler, Chen and Jiang, Yifeng and Coumans, Erwin and Luo, Zhengyi and Peng, Xue Bin and Chechik, Gal},
  booktitle={SIGGRAPH Asia},
  year={2025}
}

@article{he2026ultra,
  title   = {{ULTRA}: Unified Multimodal Control for Autonomous Humanoid Whole-Body Loco-Manipulation},
  author  = {He, Xialin and Xu, Sirui and Li, Xinyao and Dong, Runpei and Bian, Liuyu and Wang, Yu-Xiong and Gui, Liang-Yan},
  journal = {arXiv preprint arXiv:2603.03279},
  year    = {2026}
}

@article{xue2025leverb,
  title   = {{LeVERB}: Humanoid Whole-Body Control with Latent Vision-Language Instruction},
  author  = {Xue, Haoru and Huang, Xiaoyu and Niu, Dantong and Liao, Qiayuan and Kragerud, Thomas and Gravdahl, Jan Tommy and Peng, Xue Bin and Shi, Guanya and Darrell, Trevor and Sreenath, Koushil and Sastry, S. Shankar},
  journal = {arXiv preprint arXiv:2506.13751},
  year    = {2025}
}

@article{ding2025humanoidvla,
  title   = {{Humanoid-VLA}: Towards Universal Humanoid Control with Visual Integration},
  author  = {Ding, Pengxiang and Ma, Jianfei and Tong, Xinyang and Zou, Binghong and Luo, Xinxin and Fan, Yiguo and Wang, Ting and Lu, Hongchao and Mo, Panzhong and Liu, Jinxin and Wang, Yuefan and Zhou, Huaicheng and Feng, Wenshuo and Liu, Jiacheng and Huang, Siteng and Wang, Donglin},
  journal = {arXiv preprint arXiv:2502.14795},
  year    = {2025}
}

@inproceedings{deng2025human,
  title     = {Human-Object Interaction via Automatically Designed {VLM}-Guided Motion Policy},
  author    = {Deng, Zekai and Shi, Ye and Ji, Kaiyang and Xu, Lan and Huang, Shaoli and Wang, Jingya},
  booktitle = {ICLR},
  year      = {2026}
}

@article{kalaria2025dreamcontrol,
  title   = {{DreamControl}: Human-Inspired Whole-Body Humanoid Control for Scene Interaction via Guided Diffusion},
  author  = {Kalaria, Dvij and Harithas, Sudarshan S. and Katara, Pushkal and Kwak, Sangkyung and Bhagat, Sarthak and Sastry, Shankar and Sridhar, Srinath and Vemprala, Sai and Kapoor, Ashish and Huang, Jonathan Chung-Kuan},
  journal = {arXiv preprint arXiv:2509.14353},
  year    = {2025}
}

@article{hansen2001cmaes,
  title   = {Completely Derandomized Self-Adaptation in Evolution Strategies},
  author  = {Hansen, Nikolaus and Ostermeier, Andreas},
  journal = {Evolutionary Computation},
  year    = {2001}
}

@inproceedings{ma2024eureka,
  title     = {Eureka: Human-Level Reward Design via Coding Large Language Models},
  author    = {Ma, Yecheng Jason and Liang, William and Wang, Guanzhi and Huang, De-An and Bastani, Osbert and Jayaraman, Dinesh and Zhu, Yuke and Fan, Linxi and Anandkumar, Anima},
  booktitle = {ICLR},
  year      = {2024}
}

@inproceedings{xie2024text2reward,
  title     = {{Text2Reward}: Reward Shaping with Language Models for Reinforcement Learning},
  author    = {Xie, Tianbao and Zhao, Siheng and Wu, Chen Henry and Liu, Yitao and Luo, Qian and Zhong, Victor and Yang, Yanchao and Yu, Tao},
  booktitle = {ICLR},
  year      = {2024}
}

@inproceedings{ma2024dreureka,
  title     = {{DrEureka}: Language Model Guided Sim-to-Real Transfer},
  author    = {Ma, Yecheng Jason and Liang, William and Wang, Hung-Ju and Zhu, Yuke and Fan, Linxi and Bastani, Osbert and Jayaraman, Dinesh},
  booktitle = {RSS},
  year      = {2024}
}

@inproceedings{zhang2025orso,
  title     = {{ORSO}: Accelerating Reward Design via Online Reward Selection and Policy Optimization},
  author    = {Zhang, Chen Bo Calvin and Hong, Zhang-Wei and Pacchiano, Aldo and Agrawal, Pulkit},
  booktitle = {ICLR},
  year      = {2025}
}

@inproceedings{yu2023l2r,
  title     = {Language to Rewards for Robotic Skill Synthesis},
  author    = {Yu, Wenhao and Gileadi, Nimrod and Fu, Chuyuan and Kirmani, Sean and Lee, Kuang-Huei and Gonzalez Arenas, Montserrat and Chiang, Hao-Tien Lewis and Erez, Tom and Hasenclever, Leonard and Humplik, Jan and Ichter, Brian and Xiao, Ted and Xu, Peng and Zeng, Andy and Zhang, Tingnan and Heess, Nicolas and Sadigh, Dorsa and Tan, Jie and Tassa, Yuval and Xia, Fei},
  booktitle = {CoRL},
  year      = {2023}
}

@inproceedings{liang2023codeaspolicies,
  title     = {Code as Policies: Language Model Programs for Embodied Control},
  author    = {Liang, Jacky and Huang, Wenlong and Xia, Fei and Xu, Peng and Hausman, Karol and Ichter, Brian and Florence, Pete and Zeng, Andy},
  booktitle = {ICRA},
  year      = {2023}
}

@inproceedings{zha2024droc,
  title     = {Distilling and Retrieving Generalizable Knowledge for Robot Manipulation via Language Corrections},
  author    = {Zha, Lihan and Cui, Yuchen and Lin, Li-Heng and Kwon, Minae and Gonzalez Arenas, Montserrat and Zeng, Andy and Xia, Fei and Sadigh, Dorsa},
  booktitle = {ICRA},
  year      = {2024}
}

@article{nvidia2026aspire,
  title   = {{ASPIRE}: Agentic Skills Discovery for Robotics},
  author  = {Lu, Runyu and Wu, Yubo and Kou, Ethan and Fu, Letian and Xiao, Wenli and Mandlekar, Ajay and Xu, Yinzhen and Shi, Guanya and Goldberg, Ken and Chen, Ang and Chowdhury, Mosharaf and Zhu, Yuke and Fan, Linxi and Wang, Guanzhi},
  journal = {arXiv preprint arXiv:2607.00272},
  year    = {2026}
}

@inproceedings{zhou2024soar,
  title     = {Autonomous Improvement of Instruction Following Skills via Foundation Models},
  author    = {Zhou, Zhiyuan and Atreya, Pranav and Lee, Abraham and Walke, Homer Rich and Mees, Oier and Levine, Sergey},
  booktitle = {CoRL},
  year      = {2024}
}

@article{elmaaroufi2026rho,
  title   = {{RHO}: Your Coding Agent is Secretly a Roboticist},
  author  = {Elmaaroufi, Karim and Svegliato, Justin and Kalade, Sarunas and Schelle, Graham and Seshia, Sanjit A. and Zaharia, Matei},
  journal = {arXiv preprint arXiv:2606.16458},
  year    = {2026}
}

@article{xiao2026enpire,
  title   = {{ENPIRE}: Agentic Robot Policy Self-Improvement in the Real World},
  author  = {Xiao, Wenli and Xie, Jia and Zhang, Tonghe and Lin, Haotian and Fu, Letian and Xue, Haoru and Lu, Jalen and Yang, Yi and Dai, Cunxi and Wang, Zi and Wu, Jimmy and Wang, Guanzhi and Sastry, S. Shankar and Goldberg, Ken and Fan, Linxi and Zhu, Yuke and Shi, Guanya},
  journal = {arXiv preprint arXiv:2606.19980},
  year    = {2026}
}

@inproceedings{srivastava2026rosetta,
  title     = {{ROSETTA}: Constructing Code-Based Reward from Unconstrained Language Preference},
  author    = {Srivastava, Sanjana and Wang, Kangrui and Chan, Yung-Chieh and Dai, Tianyuan and Li, Manling and Zhang, Ruohan and Xu, Mengdi and Wu, Jiajun and Fei-Fei, Li},
  booktitle = {ICLR},
  year      = {2026}
}

@inproceedings{gao2025rfagent,
  title     = {{RF-Agent}: Automated Reward Function Design via Language Agent Tree Search},
  author    = {Gao, Ning and Zhang, Xiuhui and Jiang, Xingyu and You, Mukang and Zhang, Mohan and Deng, Yue},
  booktitle = {NeurIPS},
  year      = {2025}
}

@article{romeraparedes2024funsearch,
  title   = {Mathematical Discoveries from Program Search with Large Language Models},
  author  = {Romera-Paredes, Bernardino and Barekatain, Mohammadamin and Novikov, Alexander and others},
  journal = {Nature},
  year    = {2024}
}

@inproceedings{jiang2026wholebodyvla,
  title     = {{WholeBodyVLA}: Towards Unified Latent {VLA} for Whole-Body Loco-Manipulation Control},
  author    = {Jiang, Haoran and Chen, Jin and Bu, Qingwen and Chen, Li and Shi, Modi and Zhang, Yanjie and Li, Delong and Suo, Chuanzhe and Wang, Chuang and Peng, Zhihui and Li, Hongyang},
  booktitle = {ICLR},
  year      = {2026}
}

@article{deepmimic,
  title={Deepmimic: Example-guided deep reinforcement learning of physics-based character skills},
  author={Peng, Xue Bin and Abbeel, Pieter and Levine, Sergey and Van de Panne, Michiel},
  journal={ACM Transactions On Graphics (TOG)},
  volume={37},
  number={4},
  pages={1--14},
  year={2018},
  publisher={ACM New York, NY, USA}
}

@article{zhao2025resmimic,
  title   = {{ResMimic}: From General Motion Tracking to Humanoid Whole-Body Loco-Manipulation via Residual Learning},
  author  = {Zhao, Siheng and Ze, Yanjie and Wang, Yue and Liu, C. Karen and Abbeel, Pieter and Shi, Guanya and Duan, Rocky},
  journal = {arXiv preprint arXiv:2510.05070},
  year    = {2025}
}

@article{xie2026grail,
  title   = {{GRAIL}: Generating Humanoid Loco-Manipulation from {3D} Assets and Video Priors},
  author  = {Xie, Tianyi and Zhang, Haotian and Park, Jinhyung and Wang, Zi and Wen, Bowen and Li, Jiefeng and Li, Xueting and Ben, Qingwei and Weng, Haoyang and Ye, Yufei and Minor, David and Wang, Tingwu and Jiang, Chenfanfu and Fidler, Sanja and Kautz, Jan and Fan, Linxi and Zhu, Yuke and Luo, Zhengyi and Iqbal, Umar and Yuan, Ye},
  journal = {arXiv preprint arXiv:2606.05160},
  year    = {2026}
}

@article{zhao2023fastsam,
  title={Fast Segment Anything},
  author={Zhao, Xu and Ding, Wenchao and An, Yongqi and Du, Yinglong and Yu, Tao and Li, Min and Tang, Ming and Wang, Jinqiao},
  journal={arXiv preprint arXiv:2306.12156},
  year={2023}
}

@inproceedings{radford2021clip,
  title={Learning Transferable Visual Models From Natural Language Supervision},
  author={Radford, Alec and Kim, Jong Wook and Hallacy, Chris and Ramesh, Aditya and Goh, Gabriel and Agarwal, Sandhini and Sastry, Girish and Askell, Amanda and Mishkin, Pamela and Clark, Jack and Krueger, Gretchen and Sutskever, Ilya},
  booktitle={ICML},
  year={2021}
}
\bibliographystyle{arxiv}

\clearpage
\appendix
\section{Motor Model}
\label{sec:app:motor}
\label{sec:app:impl}

This section details the object-aware motor model of Sec.~\ref{sec:method:fb}, which is pretrained once before any reward-program search and stays fixed during search.

\subsection{Interaction features}
\label{sec:app:features}

For the G1 with rubber hands, the 105-dimensional interaction encoding tells the controller where the object is and how close each body link is to its surface. It concatenates a 15-dimensional object state with a 90-dimensional link term: one 3-D displacement from each of 30 body links to the object surface. The object state holds position, a six-dimensional orientation, and linear and angular velocity. Horizontal position is relative to the robot root and height stays in world coordinates, while orientation and velocities use the heading frame. For each link, the displacement to the nearest sampled object-surface point is normalized and weighted by an exponential of its distance. These features encode proximity rather than contact force.
The Inspire model keeps the same object state and adds the 24 Inspire finger links (12 per hand) to the link geometry, 54 links in total.

\subsection{Residual architecture and training objective}
\label{sec:app:residual}

The actor, $F$, and $B$ each keep their body branch frozen and train only their object residual (Eq.~\ref{eq:residual}) on the human-object interaction data of Sec.~\ref{sec:exp:setup}, so that training starts from the pretrained body prior.

\noindent\textbf{Actor objective.}
FB learning estimates discounted occupancies from replay transitions, a latent-conditioned discriminator keeps motion close to the demonstrations, and an auxiliary critic carries stability regularizers~\citep{tirinzoni2025motivo,li2026bfmzero}. The actor maximizes the FB task value $F(\boldsymbol s,\boldsymbol a,\boldsymbol z)^\top\boldsymbol z$ together with the discriminator critic $Q_D$ and the auxiliary critic $Q_A$:
\begin{equation}
 \mathcal L_\pi =
 -\mathbb E\!\left[
 F(\boldsymbol s,\boldsymbol a,\boldsymbol z)^\top \boldsymbol z+
 \lambda_D Q_D(\boldsymbol s,\boldsymbol a,\boldsymbol z)+\lambda_A Q_A(\boldsymbol s,\boldsymbol a,\boldsymbol z)
 \right],\qquad \boldsymbol a=\pi(\boldsymbol o,\boldsymbol z).
 \label{eq:app:actor}
\end{equation}
The effective weights $\lambda_D$ and $\lambda_A$ are nominal coefficients multiplied by the detached mean absolute FB value, so regularization scales with the learned task-value estimates.

\noindent\textbf{Imitation and auxiliary rewards.}
The imitation reward behind $Q_D$ is the log-odds of the discriminator probability $D(\boldsymbol s,\boldsymbol z)$,
\begin{equation}
 r_D(\boldsymbol s,\boldsymbol z)=\log D(\boldsymbol s,\boldsymbol z)-\log(1-D(\boldsymbol s,\boldsymbol z)),
 \label{eq:app:style}
\end{equation}
with probabilities clipped for numerical stability. Expert latents are backward embeddings of demonstration windows, and the discriminator is trained as a conditional classifier with gradient regularization. Auxiliary rewards penalize undesirable motion and physical violations, and $Q_A$ is trained on their normalized return. Both scalar critics and the FB learner use target networks.

\subsection{Embodiment and hand control}
\label{sec:app:embodiment}

The actor produces actions for 29 body joints, which are scaled into joint-position targets for low-level control. In actuated-hand embodiments, a separate finger controller supplies finger actuation. For the Inspire hands on the G1, a low-dimensional grasp interface replaces free exploration of all hand actuation: a per-hand scalar that the policy outputs next to its 29 body actions, from the same observation and prompt $\boldsymbol z$, is expanded to authored finger targets through software mimic coupling.

\subsection{Motor-model diagnostics}
\label{sec:app:motor-evaluation}
\label{sec:exp:motor-design}

Two ablations test the choices of Sec.~\ref{sec:app:features}--\ref{sec:app:residual}. Each variant is retrained on one GPU with a smaller training budget than the main model and evaluated with the tracking protocol of Sec.~\ref{sec:app:tracking}. The main-model rows therefore serve as a reference rather than a matched control.

\noindent\textbf{Interaction inputs.}
Table~\ref{tab:control:features} removes the object state or the link geometry from the 105-dimensional encoding, which changes only what the controller observes while the verifier's state stays intact. Without link geometry, body error is the lowest of all variants, yet object error stays at the level of the body-only BFM-Zero in Table~\ref{tab:tracking:refinement}. Without object state, object error is lower than without geometry, but success still falls to about a third of the full model's.

\begin{table}[ht]
\centering
\small
\caption{\textbf{Interaction observations.} Ablations remove the object state or the link geometry. $\Delta$SR is relative to the full 105-dimensional encoding, which is the main model of Table~\ref{tab:tracking:refinement} (without replanning).}
\label{tab:control:features}
\setlength{\tabcolsep}{5pt}
\begin{tabular}{@{}lrcccc@{}}
\toprule
Interaction encoding & Dim. & $E_h\downarrow$ & $E_o\downarrow$ & SR$\uparrow$ & $\Delta$SR \\
\midrule
\rowcolor{gray!10}
\textbf{Object state + link geometry} & 105 & 27.20 & 30.80 & 60 & \notapplicable \\
No object state & 90 & 28.57 & 49.20 & 20 & $-40$ \\
No link geometry & 15 & 25.05 & 60.83 & 12 & $-48$ \\
\bottomrule
\end{tabular}
\end{table}

\noindent\textbf{Residual capacity.}
Table~\ref{tab:control:capacity} varies the residual branch from small MLPs to full width, with shared training data and body prior. The from-scratch model is trained on the same interaction data without the body prior.

\begin{table}[ht]
\centering
\small
\caption{\textbf{Residual capacity.} MLP width$\times$depth of the residual branch; the full-width
residual mirrors the prior's module architectures and is the main model of Table~\ref{tab:tracking:refinement} (without replanning). Parameter counts include all trainable modules.
Errors and SR follow Table~\ref{tab:tracking:refinement}.}
\label{tab:control:capacity}
\vspace{8pt}
\setlength{\tabcolsep}{5pt}
\begin{tabular}{@{}lrccc@{}}
\toprule
Model & \shortstack{Trainable\\parameters} & $E_h\downarrow$ & $E_o\downarrow$ & SR$\uparrow$ \\
\midrule
From scratch & 843\,M & 60.96 & 78.21 & 2 \\
\midrule
\multicolumn{5}{@{}l}{\textit{Residual on the body prior}} \\
MLP $256\times$$2$ & 14.1\,M & 36.19 & 50.81 & 28 \\
MLP $512\times2$ & 30.5\,M & 31.51 & 42.58 & 30 \\
MLP $1024\times2$ & 79.5\,M & 29.22 & 37.42 & 46 \\
\rowcolor{gray!10}
\textbf{Full-width residual} & 852.0\,M & 27.20 & 30.80 & 60 \\
\bottomrule
\end{tabular}
\end{table}

\section{Reward Inference}
\label{sec:app:inference}

Every prompt the motor model executes is computed through $B$, either from a reward scored over the bank or directly from reference states.

\subsection{Uniform and reward-tilted aggregation}
\label{sec:app:aggregation}

The FB factorization gives an approximate value for any integrable state reward:
\begin{equation}
\begin{aligned}
Q^{\pi_{\boldsymbol z}}_r(\boldsymbol s,\boldsymbol a)
 &= \int r(\boldsymbol s')\,M^{\pi_{\boldsymbol z}}(\mathrm{d}\boldsymbol s'\mid \boldsymbol s,\boldsymbol a)\\
 &\approx F(\boldsymbol s,\boldsymbol a,\boldsymbol z)^{\top}\,
       \mathbb E_{\boldsymbol s'\sim\rho}[r(\boldsymbol s')B(\boldsymbol s')].
\end{aligned}
\label{eq:app:value}
\end{equation}
The expectation is the prompt $\widetilde{\boldsymbol z}_r$ of Sec.~\ref{sec:method:fb}, which we estimate on the reward-inference bank $\mathcal B$.

\noindent\textbf{Bank construction.}
We build $\mathcal B$ once, after motor-model pretraining and before agent search. Its body-object states come from the source distribution used to learn the FB representation: replay from multi-task, multi-object loco-manipulation training. We add no separate rollouts: the replay already contains the reference-following rollouts collected during training, which cover interaction phases such as contact acquisition, transport, and release. For each state we cache $B(\boldsymbol s_n)$ and the reward-feature tensors that programs read (namespace \texttt{F} in Sec.~\ref{sec:app:program}). The bank stays fixed during evolution. Target-task rollouts update the skill library $\mathcal H$ but add no states to $\mathcal B$, and only the bank diagnostics of Sec.~\ref{sec:app:fb-ablations} change it.

\noindent\textbf{Uniform aggregation.}
Averaging $r(\boldsymbol s_n)B(\boldsymbol s_n)$ over $\mathcal B$ estimates the expectation in Eq.~\ref{eq:app:value}. The projection $\operatorname{proj}$ then rescales the nonzero result to the sphere radius used in pretraining, block by block when the latent has separately normalized blocks. Scaling the whole reward by a positive constant therefore leaves the prompt unchanged, so only relative reward weights matter.

\noindent\textbf{Reward-tilted aggregation.}
Reward tilting shifts the estimate toward high-reward bank states:
\begin{equation}
 w_n=\frac{\exp(\beta r_n)}{\sum_{k=1}^{N}\exp(\beta r_k)},\qquad
 \boldsymbol z_r=\operatorname{proj}\!\left(\sum_{n=1}^{N}w_n r_n B(\boldsymbol s_n)\right),
 \quad r_n=r(\boldsymbol s_n),\quad \beta\geq 0.
 \label{eq:app:weighted}
\end{equation}
With the stage rewards $r_n=\hat r_{j,n,t}$, Eq.~\ref{eq:app:weighted} gives the prompt $\boldsymbol z_{j,t}$ of Eq.~\ref{eq:zr} with $\alpha_{j,n,t}\propto w_n r_n$, because $\operatorname{proj}$ removes the overall scale. At $\beta=0$ it reduces to uniform aggregation, and larger $\beta$ concentrates the estimate on high-reward bank states. The tilt is therefore a biased heuristic. Reward scale and $\beta$ jointly set its concentration, and a sharply concentrated weighting makes $\boldsymbol z_r$ depend on a few bank states.

\subsection{Object-anchored drift correction}
\label{sec:app:drift-correction}

The actor observes body and object only in local frames (Sec.~\ref{sec:app:features}), so it cannot see how far execution has drifted from the reference in the world frame. The world-frame position error of an anchor (the root, the object, or both) gives a proportional velocity correction with gain $k_p$. The correction is bounded, added to the anchor's reference velocity, and expressed in the live heading frame. A Gaussian reward scores each bank state's anchor velocity against this command. We center the reward over $\mathcal B$ and infer a correction latent, which is normalized separately, blended with the native reference latent $\boldsymbol z_t^{\mathrm{ref}}$, the projected mean of $B$ over a short window of upcoming reference states, by a relative coefficient, and projected back onto the latent sphere. An optional threshold gain $\gamma$ scales the gain to $k_p\,(1+\gamma\,(\lVert\Delta \boldsymbol p_t\rVert/0.5)^2)$, raising it as the offset $\Delta\boldsymbol p_t$ approaches the 0.5\,m termination distance.

\begin{table}[t]
\centering
\small
\caption{\textbf{Drift-correction ablation.} Tracking protocol of Table~\ref{tab:tracking:refinement}; $E_h$ and $E_o$ in cm. The anchor is the quantity steered toward the reference, $k_p$ is the correction gain, and the shaded row is the configuration used in Table~\ref{tab:tracking:refinement}.}
\label{tab:app:drift-ablation}
\setlength{\tabcolsep}{5pt}
\begin{tabular}{@{}llcccc@{}}
\toprule
Anchor & Addition & $k_p$ & SR (\%)$\uparrow$ & $E_h\downarrow$ & $E_o\downarrow$ \\
\midrule
None & -- & -- & 60 & 27.20 & 30.80 \\
\midrule
\multicolumn{6}{@{}l}{\textit{Anchor}} \\
Root & -- & 2 & 68 & 19.56 & 25.27 \\
\rowcolor{gray!10}
\textbf{Object} & -- & 3 & \textbf{72} & 19.80 & 24.92 \\
Root and object & -- & 2, 3 & 68 & \textbf{18.24} & \textbf{24.10} \\
\midrule
\multicolumn{6}{@{}l}{\textit{Correction gain $k_p$}} \\
Root & -- & 1 & 62 & 19.76 & 25.40 \\
Root & -- & 3 & 64 & 19.44 & 24.94 \\
Object & -- & 2 & 64 & 20.57 & 25.77 \\
Object & -- & 4 & 68 & 19.99 & 24.69 \\
Object & -- & 5 & 70 & 19.77 & 24.84 \\
\midrule
\multicolumn{6}{@{}l}{\textit{Design choices}} \\
Root & Update every 2 steps & 2 & 66 & 19.55 & 25.23 \\
Root & Kernel width 0.5\,m/s & 2 & 64 & 18.82 & 24.43 \\
Object & Kernel width 0.5\,m/s & 3 & 68 & 19.57 & 25.05 \\
\midrule
\multicolumn{6}{@{}l}{\textit{Additions to the object anchor}} \\
Object & Derivative term, $k_d=2$ & 3 & 62 & 20.33 & 25.56 \\
Object & Derivative term, $k_d=5$ & 3 & 66 & 20.38 & 25.26 \\
Object & Object-yaw feedback, gain 0.5 & 3 & 70 & 20.74 & 25.69 \\
Object & Object-yaw feedback, gain 1.5 & 3 & 66 & 22.12 & 27.00 \\
Object & Threshold gain, $\gamma=1$ & 3 & 66 & 19.85 & 24.72 \\
Object & Threshold gain, $\gamma=3$ & 3 & 68 & 19.96 & 25.08 \\
\bottomrule
\end{tabular}
\end{table}

\begin{figure}[t]
\centering
\begin{minipage}[t]{0.49\linewidth}\centering
\includegraphics[width=\linewidth]{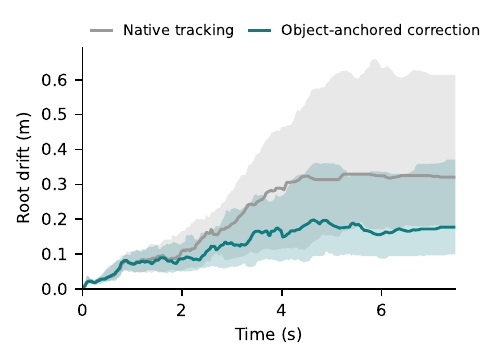}
\end{minipage}\hfill
\begin{minipage}[t]{0.49\linewidth}\centering
\includegraphics[width=\linewidth]{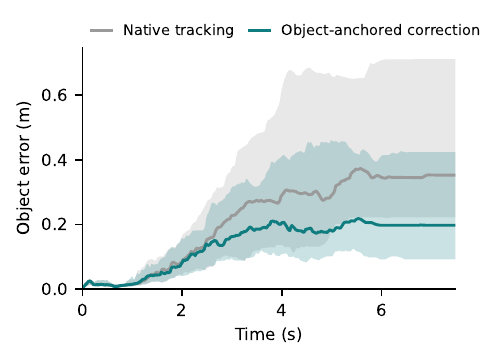}
\end{minipage}
\caption{\textbf{World-frame drift over time.} Horizontal distance between the simulated and reference root (left) and object (right) for native tracking and the object-anchored drift correction, on the 50 largebox clips of Table~\ref{tab:tracking:refinement}. Lines show the median and bands the 25--75\% range over all 50 motions; each motion is counted to the end of its clip, and after its clip ends its final value is held. Without correction the median root drift grows to 0.29\,m at 4\,s and 0.32\,m by the end, against 0.16\,m and 0.18\,m with the correction, which levels off after about 4\,s. The two curves coincide for the first second because the correction has a 0.10\,m deadband.}
\label{fig:app:drift}
\end{figure}

\noindent\textbf{Anchor, gain, and additions.}
Table~\ref{tab:app:drift-ablation} shows that every variant improves on tracking without correction. Each anchor has its own best gain, $k_p=2$ for the root and $k_p=3$ for the object, because moving the object requires moving the body first. Anchoring both root and object gives the lowest errors but a lower SR than the plain object anchor, which Table~\ref{tab:tracking:refinement} therefore uses.

\subsection{\texorpdfstring{Bank size, reward tilt, and cost}{Bank size, reward tilt, and cost}}
\label{sec:app:fb-ablations}
\label{sec:exp:coverage}

The bank determines which behaviors a reward can reach and how much each projection costs. Bank studies therefore change only $\mathcal B$, keeping the controller, reward program, skill library, and test scenarios fixed.

We run these studies on the feet-only task, whose target behavior is rare in the replay. Every bank is therefore a random draw from a pool that over-samples states showing this behavior, and count comparisons draw nested banks of different sizes from it. Tilt comparisons share the bank, reward, and checkpoint, vary $\beta$ with reward scale fixed, and run without agent evolution. Table~\ref{tab:app:fb-bank} pairs success with $N_{90}$, the number of states that carry 90\% of the weight, and with projection latency and peak memory. Latency covers repeated reward projection on identical hardware and excludes one-time bank construction and feature caching.

\begin{figure}[t]
\centering
\includegraphics[width=\linewidth]{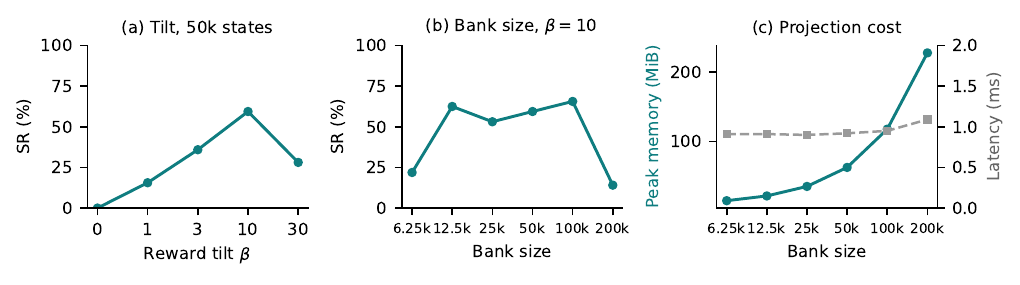}
\caption{\textbf{Reward tilt, bank size, and cost on the feet-only task.} (a) SR against the reward tilt $\beta$ with 50k states. (b) SR against bank size at $\beta=10$. (c) Peak memory (solid) and projection latency (dashed) against bank size.}
\label{fig:app:fb-scaling}
\end{figure}

\noindent\textbf{Where the weight goes.}
$N_{90}$ explains both failure modes summarized in Sec.~\ref{sec:exp:skills}. Uniform weights spread over the whole bank, and $\beta=30$ puts nearly all weight on three states. At $\beta=10$ the weight moves with the context: at most frames it sits on about ten states, and during contact it briefly spreads over thousands. The 200k bank fails because its tilted weight concentrates on an order of magnitude fewer states than the 100k bank. Peak memory grows linearly with bank size (Fig.~\ref{fig:app:fb-scaling}c).

\begin{table}[ht]
\centering
\small
\caption{\textbf{Reward-inference bank diagnostics on the feet-only task.} SR is the task success rate of the fixed program; $N_{90}$ is the number of bank states that carry 90\% of the weight at a frame, averaged over the frames of a rollout and reported as the median over the 64 test scenarios. Latency is per reward projection; peak memory covers resident bank features and projection workspace.}
\label{tab:app:fb-bank}
\setlength{\tabcolsep}{4pt}
\begin{tabular}{@{}lccccc@{}}
\toprule
$\beta$ & \shortstack{State\\count} & \shortstack{SR\\(\%)$\uparrow$} & $N_{90}$ & \shortstack{Projection\\latency (ms)$\downarrow$} & \shortstack{Peak\\memory (MiB)$\downarrow$} \\
\midrule
\multicolumn{6}{@{}l}{\textit{Reward tilt}} \\
0 (uniform) & 50k & 0.0 & 45{,}000 & -- & -- \\
1 & 50k & 15.6 & 28{,}674 & -- & -- \\
3 & 50k & 35.9 & 10{,}655 & -- & -- \\
\rowcolor{gray!10}
10 & 50k & \textbf{59.4} & 2{,}077 & 0.92 & 63 \\
30 & 50k & 28.1 & 3 & -- & -- \\
\midrule
\multicolumn{6}{@{}l}{\textit{Bank size}} \\
10 & 6.25k & 21.9 & 293 & 0.91 & 15 \\
10 & 12.5k & 62.5 & 346 & 0.91 & 22 \\
10 & 25k & 53.1 & 1{,}441 & 0.90 & 36 \\
10 & 100k & \textbf{65.6} & 2{,}815 & 0.95 & 118 \\
10 & 200k & 14.1 & 170 & 1.09 & 228 \\
\bottomrule
\end{tabular}
\end{table}

\section{Reward Programs and the Evolution Loop}
\label{sec:app:evolution}

This section makes the programs and the search loop of Algorithm~\ref{alg:search} concrete.
\begin{algorithm}[t]
\caption{Reward-program evolution at test time.}
\label{alg:search}
\begin{algbox}
\small
\noindent\textbf{Input:} task $\ell$, scene $c$, verifier $\mathcal C$, motor model $\pi$ with bank $\mathcal B$, skill library $\mathcal H$, round budget $K$.
\begin{pseudo}
  \item The agent writes an initial program from $\ell$, $c$, $\mathcal C$, and $\mathcal H$; evaluate it and make it the current best.
  \item \algkw{for} $k=1,\ldots,K$ \algkw{do}
  \item \algind Prompt the agent with $\ell$, $c$, $\mathcal C$, $\mathcal H$, and the current best program with its rollout feedback.
  \item \algind Parse and validate the proposed programs; discard invalid candidates.
  \item \algind \textit{Inner loop:} tune the constants $\boldsymbol{\theta}$ of each valid program with CMA-ES on the search scenarios.
  \item \algind Compare the tuned candidates with the current best on the same search scenarios.
  \item \algind Re-evaluate the top candidate and the current best on the confirmation scenarios; the candidate becomes the current best if its gain exceeds evaluation noise.
  \item \algind Summarize criterion- and stage-level feedback.
  \item \algind \algkw{if} the current best satisfies every criterion in $\mathcal C$ \algkw{then break}
  \item \algkw{end for}
  \item Add the current best program and its verifier results to $\mathcal H$.
\end{pseudo}
\noindent\textbf{Output:} the current best program.
\end{algbox}
\end{algorithm}
\subsection{Program semantics}
\label{sec:app:program}

A program $P=\big(\boldsymbol{\theta},\{(r_j,g_j)\}_{j=1}^{J}\big)$ exposes each part of a strategy as a separately editable component (Table~\ref{tab:app:program}): tunable constants and, per stage, reward code, a completion condition, and optional local memory. Completion conditions make temporal structure explicit: a stage can persist until $g_j$ holds, enforce a dwell time, time out, or return to an earlier stage.

\begin{table}[ht]
\centering
\small
\caption{\textbf{Editable components of a reward program.} Each edit is judged by its executed outcome.}
\label{tab:app:program}
\setlength{\tabcolsep}{4pt}
\begin{tabular}{@{}llll@{}}
\toprule
Component & Symbol & Role & Example edit \\
\midrule
Reward code & $r_j$ & Desired body-object interaction & Upward $\to$ horizontal object motion \\
Constants & $\boldsymbol\theta$ & Scale and target of an objective & Target height, contact offset \\
Stages & $j$ & Intermediate objectives & Split acquisition from transport \\
Completion & $g_j$ & Completing or switching a stage & Require support before transport \\
Stage memory & -- & Context captured at runtime & Displacement since stage entry \\
\bottomrule
\end{tabular}
\end{table}

\noindent\textbf{Bank features and live context.}
Program code reads bank features and live context from separate namespaces, so that a reward means the same thing over the bank as in the rollout that executes its latent. \texttt{F} holds bank features, each a tensor over all $N$ bank states, so one call to \texttt{reward} returns the stage rewards $\hat r_{j,n,t}$ of Eq.~\ref{eq:zr} for the whole bank at once. \texttt{C} holds live context: scalars read from the current frame, such as the object-target distance, and the unit direction \texttt{C["v\_dir"]} from the object toward its target. The split is needed because a live measurement is not a property of the bank states. A current contact report, for instance, says nothing about contact forces in a bank that stores only geometry. \texttt{W} exposes the constants $\boldsymbol\theta$, namely relative reward weights and thresholds or kernel widths, and expressions can read only this declared interface.

\noindent\textbf{A complete program.}
The selected program for \texttt{lift\_rel} (place the box on a 0.65\,m table and let go) has two stages and seven constants. In code, $r_j$ is the function \texttt{reward} and $g_j$ is \texttt{subgoal}:
{\scriptsize
\begin{verbatim}
name    : lift_carry_release_2stage
weights : [1.460, 0.440, 0.212, 0.203, 1.578, 0.310, 1.961]
           lift   drive  damp   arrive release still  settle

stage 0  # carry, and place
  def reward(F, C, W):
      high  =  W[0] * sigmoid((F["obj_z"] - 0.85)/0.05) * F["hand_contact"]
      drive =  W[1] * (F["obj_vel"] * C["v_dir"]).sum(-1)
      damp  = -W[2] * F["vel_norm"]
      return high + drive + damp
  def subgoal(C, W):
      return C["d_target"] < W[3]

stage 1  # release
  def reward(F, C, W):
      return ( W[4] * (1.0 - F["hand_contact"])
             - W[5] * F["vel_norm"]
             - W[6] * norm(F["root_vel_local"]) )
  def subgoal(C, W):
      return None
\end{verbatim}
}
\noindent Stage~0 ends once the live object-target distance \texttt{C["d\_target"]} falls below \texttt{W[3]}, and stage~1 then rewards release.

\subsection{Agent context}
\label{sec:app:context}

Each round, the agent receives one plain-text prompt. Every task and run uses the same template: the fixed parts below are quoted from it, lightly abridged, and \ph{placeholders} mark what each run fills in. The parts appear in the prompt in the order shown.

\noindent\textbf{Role and mechanism.} The prompt first explains how a program becomes behavior, so that the agent writes rewards for the bank rather than for a live robot.
\begin{promptbox}{Prompt: role}
You design a \textbf{staged} reward that teaches a \textbf{frozen} humanoid policy a manipulation skill. A skill is a sequence of stages. Each stage is a single behavior latent $z$ for the frozen policy, obtained by scoring the policy's $\sim$50k-state behavior buffer with a reward and taking a softmax-weighted mean. Each stage has \pc{reward(F, C, W)}, a tensor of shape $(N,)$ over the $N$ buffer states (higher = pick it), and \pc{transition(C, W)}, a bool; when true, advance to the next stage (last stage: null). \pc{W} is a numeric weight vector; CMA-ES tunes \pc{W} afterwards.
\end{promptbox}

\noindent\textbf{Feature vocabulary.} Programs can read only these names. \pc{F} is evaluated over the bank and \pc{C} on the live frame (Sec.~\ref{sec:app:program}).
\begin{promptbox}{Prompt: feature vocabulary (excerpt)}
\textbf{F[...] are per-buffer-state features} (tensors of shape $(N,)$ unless noted), egocentric:\\[1pt]
\begin{tabular}{@{}p{0.3\linewidth}p{0.64\linewidth}@{}}
\pc{F["obj_z"]} & object height (m)\\
\pc{F["obj_vel"]} $(N,3)$ & object velocity, heading frame\\
\pc{F["vel_norm"]} & object speed\\
\pc{F["hand_contact"]} & $\exp(-5\,d_{\text{hand,object}})$; high when a hand is on the object\\
\pc{F["foot_contact"]} & foot-object contact\\
\pc{F["root_vel_local"]} $(N,3)$ & robot root velocity, heading frame\\
\pc{F["robot_yaw_rate"]} & robot turning rate about the vertical (signed)\\
\end{tabular}\\[2pt]
\textbf{C[...] are live per-frame scalars} (Python floats unless noted):\\[1pt]
\begin{tabular}{@{}p{0.3\linewidth}p{0.64\linewidth}@{}}
\pc{C["d_target"]} & object-to-target distance (m)\\
\pc{C["v_dir"]} $(3,)$ & unit direction toward the target, heading frame\\
\pc{C["face"]} & how much the robot already faces the target: $+1$ facing, $0$ sideways, $-1$ behind\\
\pc{C["obj_z"]}, \pc{C["rest_z"]} & live and resting object height\\
\pc{C["d_hand"]} & hand-to-object distance (m)\\
\pc{C["progress"]} & episode progress, $0$ to $1$\\
\end{tabular}\\[1pt]
\ph{further features, and a variant for tasks without a target position}
\end{promptbox}

\noindent\textbf{Code rules.} Programs that violate these rules are rejected by the validator before any rollout.
\begin{promptbox}{Prompt: code rules (excerpt)}
The reward must reference \pc{F[...]}. Keep 1--3 stages and few weights. \pc{torch} and \pc{math} are available.
\begin{itemize}[leftmargin=1.2em,itemsep=0pt,topsep=1pt]
\item \pc{reward(F,C,W)} must return a tensor of shape $(N,)$ built from \pc{F[...]}, never a Python scalar. \pc{C[...]} entries are Python floats: never call \pc{torch.exp} or \pc{torch.sigmoid} on them.
\item \pc{C["v_dir"]} is a length-3 tensor; use \pc{C["v_dir"].unsqueeze(0)}.
\item \pc{transition(C,W)} returns a Python bool from \pc{C} and \pc{W}, e.g.\ \pc{return C["d_target"] < W[3]}; it must not read \pc{F[...]}. The last stage's transition is null.
\end{itemize}
\end{promptbox}

\noindent\textbf{Task and verifier.} The task text is followed by the verifier, one row per criterion with its definition, so feedback can refer to criteria by name.
\begin{promptbox}{Prompt: task and criteria}
\textbf{TASK:} \ph{task request and scene description}\\[2pt]
\textbf{What each tier measures.} These are definitions, not instructions: deciding which reward term satisfies each one is your job, and the right term differs by task.\\
\hspace*{1em}\ph{metric} \ph{op} \ph{threshold}\quad \ph{definition}, for example:\\
\hspace*{1em}\pc{final_hand_d > 0.35}\quad hand-to-object distance on the final frame, in metres; the only metric that separates ``placed it and let go'' from ``carried it to the mark and froze there still holding it''.\\
\hspace*{1em}\pc{final_d < 0.3}\quad 3-D distance from the object's last frame to the target, in metres.
\end{promptbox}

\noindent\textbf{Search phase.} One instruction per round sets how far a proposal may depart from the current best program.
\begin{promptbox}{Prompt: phase instruction (one per round)}
\textbf{Explore} (early rounds): vary the decomposition the most: try 1 vs.\ 2 vs.\ 3 stages, different transition conditions, and different sub-goals per stage.\\
\textbf{Refine} (middle rounds): keep the winning decomposition, but vary the reward terms inside each stage: which features, which combinations, add, drop, or reweight a term.\\
\textbf{Tune} (late rounds): keep the winning structure almost fixed and propose small variations; CMA-ES fine-tunes the numbers.
\end{promptbox}

\noindent\textbf{Reference programs with per-criterion evidence.}
Each carried program is listed with how many of the 192 confirmation rollouts pass each criterion of $\mathcal C$, followed by the full source of the three strongest programs:
\begin{promptbox}{Prompt: reference programs}
\textbf{Reference structures so far}, each showing how far up the criteria it climbed and where it is stuck. Beat the leader: graft the mechanism that got the furthest-climbing one past its criteria, then fix its stuck criterion.\\[1pt]
-- \ph{program}: tiers=\ph{m}/\ph{n}\quad stuck@\ph{criterion} (\ph{k}/192 envs clear it, median \ph{value})\\
\hspace*{1em}\ph{criterion} \ph{k}/192 \textsc{pass}/\textsc{fail}\quad \ph{criterion} \ph{k}/192 \textsc{pass}/\textsc{fail}\quad \dots\\[1pt]
\textbf{Full source of the strongest references} (graft from these, do not re-derive): \ph{three programs}
\end{promptbox}

\noindent\textbf{Which reference leads on which criterion.}
Because $\boldsymbol z$ is inferred per stage, a stage that earns a criterion can be transplanted, and the prompt says where each criterion is earned:
\begin{promptbox}{Prompt: which reference leads}
\ph{program A} leads on: \ph{criterion}=\ph{pass rate}, \dots\qquad \ph{program B} leads on: \ph{criterion}=\ph{pass rate}, \dots\\[1pt]
No single reference leads everywhere. A stage that achieves a tier can be carried over as a stage: take the stage that earns one tier from one reference and the stage that earns another from a second reference, and write the transition between them. At least one of your proposals \textbf{must} be such a merge, quoting the two sources.
\end{promptbox}

\noindent\textbf{Rollout feedback.}
The failing criterion is reported with its margin and spread, followed by a stage-resolved trace and the problems read off it:
\begin{promptbox}{Prompt: rollout feedback}
round \ph{r}: best `\ph{program}' passes \ph{m}/\ph{n} tiers (\ph{J} stages). \textbf{Stuck on tier} `\ph{criterion}' (\ph{k} of 192 envs clear it). The median env is \ph{value}, which \ph{clears | misses} the bar by \ph{margin}, so this is a \ph{tail/variance | low-mean} failure. \dots\ Find the stage that was supposed to move `\ph{criterion}' and did not, and change only that stage.\\[1pt]
\textbf{Stage trace} (median over envs):\\
\hspace*{1em}stage \ph{j}: \ph{k}/192 envs, dwell \ph{s}\,s, obj\_z \ph{a}$\to$\ph{b}, d\_target \ph{a}$\to$\ph{b}, grip \ph{a}$\to$\ph{b}\\
\textbf{Stage problems:} \ph{e.g.\ stage 0 lasted only 0.66\,s of 14.0\,s: its transition fires almost immediately}
\end{promptbox}

\noindent\textbf{Output format.}
The agent returns a JSON list of programs (Sec.~\ref{sec:app:program}). Each stage's \texttt{transition} field holds $g_j$, and \texttt{w\_init}, \texttt{w\_lo}, and \texttt{w\_hi} give initial values and bounds for $\boldsymbol\theta$:
\begin{promptbox}{Prompt: request and output format}
Propose \ph{K} diverse staged structures as the JSON list. Return strict JSON, a list of structures, each:\\
\texttt{\{"name": ..., "w\_names": [...], "w\_init": [...], "w\_lo": [...], "w\_hi": [...],}\\
\texttt{\ "stages": [\{"reward": "def reward(F,C,W): ...", "transition": "def transition(C,W): ..."\}, ...]\}}\\
The last stage's transition is null. Make the structures diverse (different numbers of stages, terms, and transitions).
\end{promptbox}

\subsection{Selection and the skill library}
\label{sec:app:selection}

Accepting an edit on the scenarios that suggested it would reward lucky edits, so search and confirmation use separate scenario grids. Candidates are developed on a 16-scenario search grid, where the current best program and each candidate share initial conditions. A 64-scenario grid then re-evaluates the pair with three repeats per scenario, and the candidate replaces the current best only if its improvement exceeds the run-to-run evaluation noise. The final evaluation re-runs the selected program on a different set of 64 scenarios drawn from the same distribution. Both grids spread the box bearing evenly over $\pm12^\circ$ and its distance over $\pm0.15$\,m around the nominal start, and every method is evaluated on the identical 64 scenarios.

The skill library $\mathcal H$ is a text document built from completed tasks. Each entry records the scene, the task text the agent saw, the verifier with the pass rate of each criterion, the measured success, and the selected program with its tuned constants. A closing section states the lessons: reward-design regularities shared across the stored programs. For example, every multi-stage program ends with a release stage that pays for $1-$contact and for object and robot stillness, and a lift term is always a product with hand contact, never a bare height term, which would select airborne states. The whole library, eight tasks in about 28\,KB, is placed in the agent's prompt every round.
\begin{figure}
  \centering
  \includegraphics[width=\linewidth]{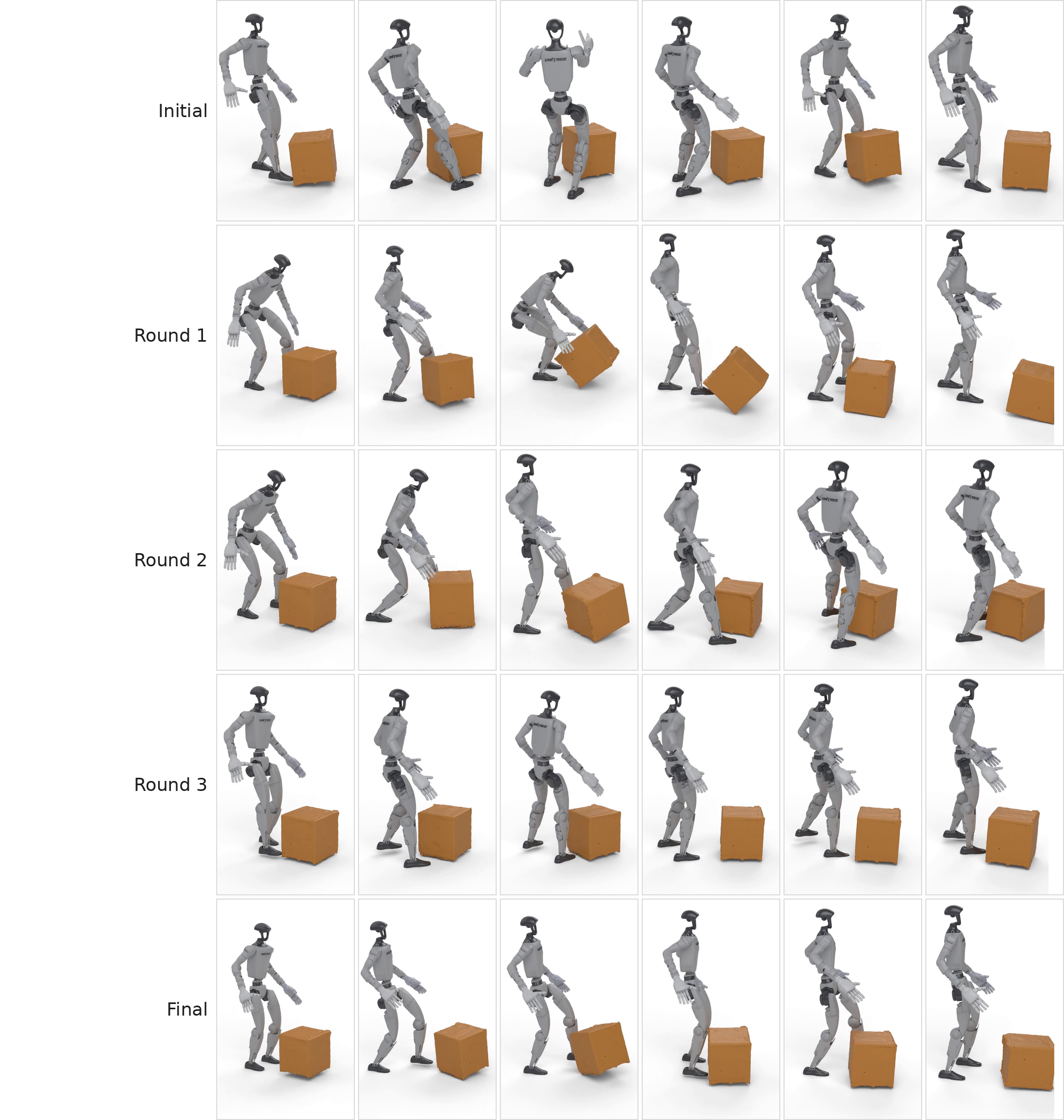}
  \caption{\textbf{Reward-program evolution on the kick task.} Each row shows the current best program after the initial proposal, rounds 1--3, and the final round (top to bottom), the kick programs evaluated in Table~\ref{tab:search:rounds}; frames run left to right. Revised objectives change the contact strategy across rounds.}
  \label{fig:evolution-trace}
\end{figure}

\begin{figure}
  \centering
  \includegraphics[width=\linewidth,height=0.9\textheight,keepaspectratio]{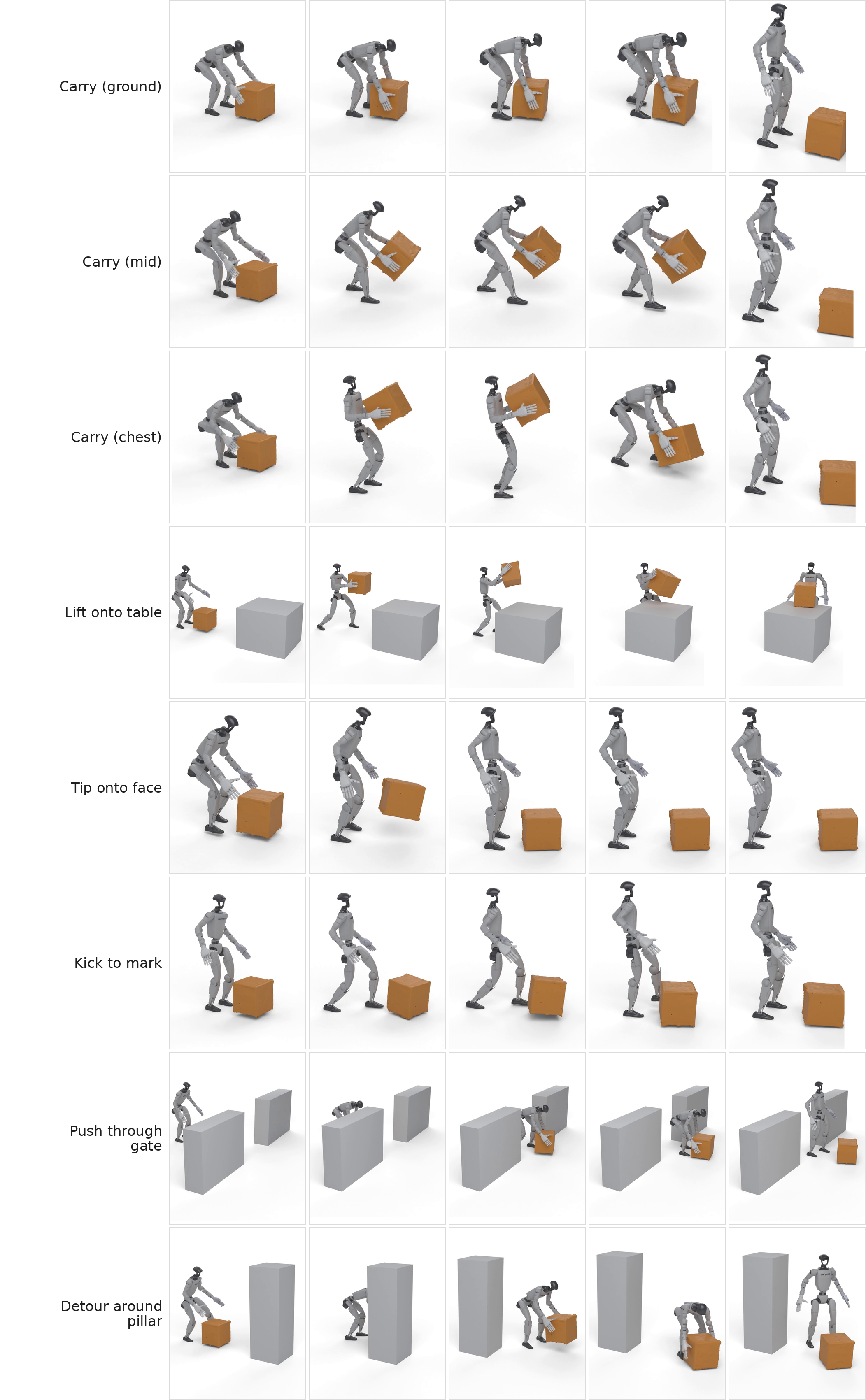}
  \caption{\textbf{The eight basic skills.} Final evolved program of each task family, frames left to right. Whole-body motion links object acquisition,
  transport, and placement across carrying, lifting, tipping, kicking, and obstacle-aware pushing.}
  \label{fig:case}
\end{figure}

\begin{figure}
  \centering
  \includegraphics[width=\linewidth]{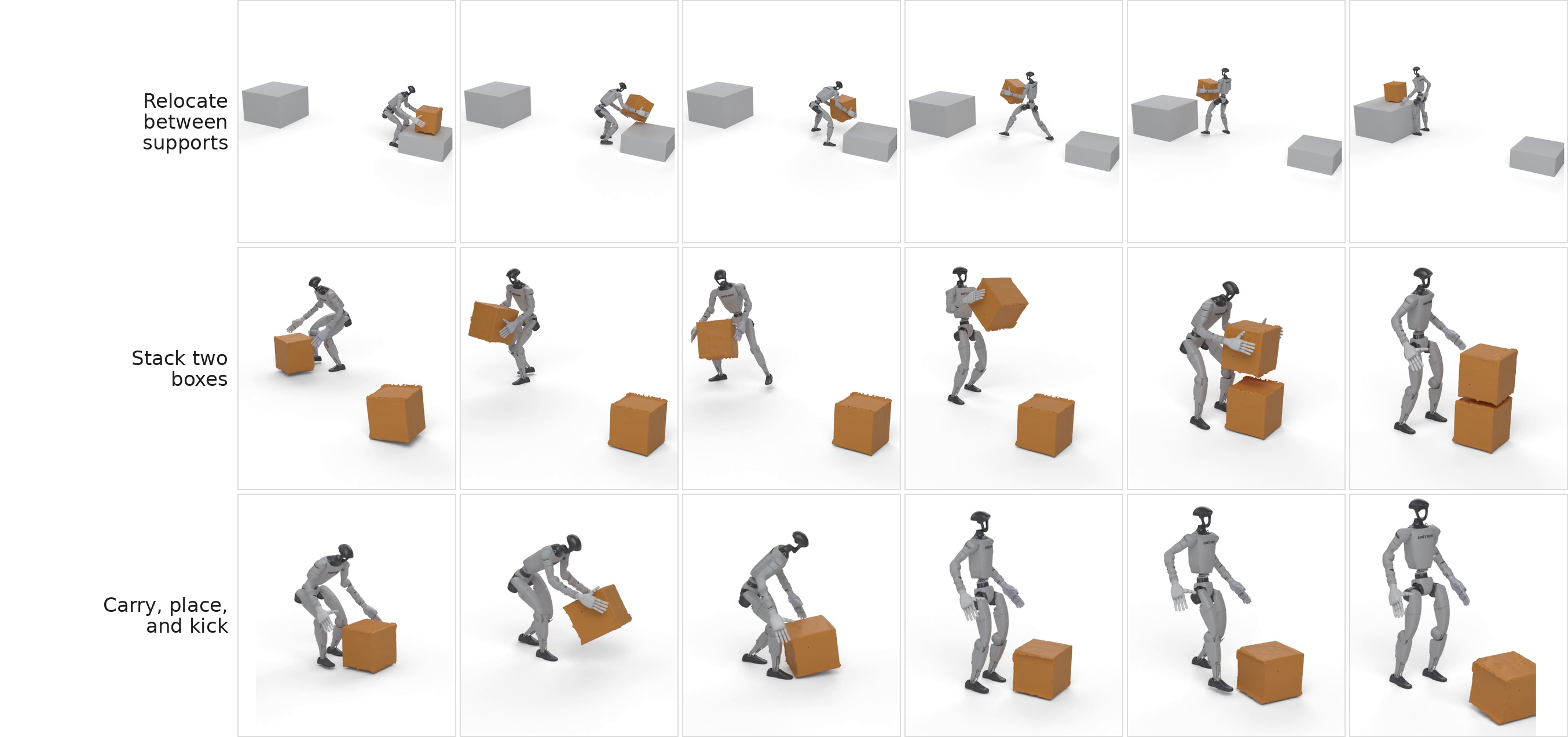}
  \caption{\textbf{Composite tasks.} The three composite tasks of Table~\ref{tab:search:library}: relocating a box between supports, stacking one box on another, and carrying, placing, then kicking a box. Frames run left to right.}
  \label{fig:push}
\end{figure}

\subsection{Evolution cost accounting}
\label{sec:app:compute}

Evolution cost counts every executed candidate, including rejected ones. CMA-ES with $G$ generations of $\lambda_{\mathrm{ES}}$ candidates on $S$ scenarios costs $\lambda_{\mathrm{ES}}GS$ rollouts per program, before confirmation. With $\lambda_{\mathrm{ES}}=4$, $G=3$, and $S=16$, this is 192 rollouts per program, and each confirmed candidate adds 64 scenarios $\times$ 3 repeats $=192$ rollouts. Every rollout runs the full episode (700 control steps at 50\,Hz), so simulator cost is proportional to the rollout count. A five-round search on one task family executes 6.7--14.2\,M environment steps, 9.5\,M on average over the eight families, or about 1.9\,M per round. Table~\ref{tab:search:comparison} reports language-model tokens. The simulator pauses while the agent revises a program.
The agent is DeepSeek-V4-Flash, queried through its API. A three-round run makes four to five calls and uses 0.13--0.18\,M tokens. Each round spends 1--2\,min in the language model and 16--34\,min in simulation on one GPU, so simulation dominates wall-clock time.

The goal-conditioned experiments of Sec.~\ref{sec:exp:skills} set budget $K=4$, so each search runs five rounds including the initial program (Table~\ref{tab:search:rounds}). Ablations that skip tuning or use a single scenario spend the saved rollouts on more proposals, which keeps their simulation budget equal to that of the full system (Sec.~\ref{sec:app:search-ablations}).
\section{Experimental Protocols and Additional Results}
\label{sec:app:eval-details}

This section gives the protocol behind each experiment of Sec.~\ref{sec:experiments}, in the same order, after the scoring and replication rules they share.

\subsection{Verifier scoring and replication}
\label{sec:app:scoring}

The verifier $\mathcal C$ is fixed apart from the candidate programs and scores every study independently of the reward being optimized. It records task outcomes, constraint satisfaction, and stage-resolved progress. Only $\mathcal C$ decides success, while program termination and reward value serve as diagnostics. Proximity, contact, lift, and sustained support are separate criteria.

\noindent\textbf{Success and earned tiers.}
Success requires every criterion in $\mathcal C$ to hold in the same rollout, and earned tiers measure partial progress. Let $\mathcal P_k$ be the criteria a candidate passes on task $k$ out of the $|\mathcal C_k|$ criteria of that task. The earned fraction is
\begin{equation}
 E_k=|\mathcal P_k|\,/\,|\mathcal C_k|.
 \label{eq:app:earned}
\end{equation}
Some criteria, such as staying upright, are already met by doing nothing. $E_k$ counts them like any other criterion, so we report an inaction reference that stands still (Table~\ref{tab:search:comparison}). The reference and all thresholds are fixed before any comparison. We aggregate success macro-averages task families after averaging test scenarios within each family.

\subsection{Reference tracking}
\label{sec:app:tracking}

Every method shares the reference, initialization, horizon, and termination protocol, and $E_h$ and $E_o$ are computed over executed frames against the time-aligned reference. An episode terminates when the root drops below $0.15$\,m, the mean object-surface error exceeds $0.5$\,m, or an interaction-consistency check fails. SR counts every start, so a trajectory that terminates early lowers SR even when its errors are small.

\begin{table}[t]
\centering
\caption{\textbf{Supplementary tracking metrics} on the clips of Table~\ref{tab:tracking:refinement}. Every metric is averaged over three evaluation runs of the 50 clips, with rates over clips rounded to integers, and the rows below the first block come from recordings in which every clip runs to its end without early termination. \emph{Orientation error}: object orientation error, averaged over the whole clip. \emph{Contact retention}: fraction of reference-contact frames in which the simulated hand is also in contact, with contact meaning a hand vertex within 3\,cm of the object surface. \emph{Slip}: wrist speed in the object frame during contact, in simulation and on the reference. \emph{Penetration}, \emph{object below ground}: frames deeper than 2\,cm. \emph{Falls}: pelvis below 0.35\,m or tilted beyond $60^\circ$.}
\label{tab:tracking_metrics}
\setlength{\tabcolsep}{4pt}
\small
\begin{tabular}{@{}l cccc@{}}
\toprule
 & ULTRA & BFM-Zero & InterEvolve & + Evolving \\
\midrule
SR (\%) $\uparrow$ & 66 & 8 & 60 & \textbf{72} \\
$E_h$ (cm) $\downarrow$ & \textbf{15.68} & 28.36 & 27.20 & 19.80 \\
$E_o$ (cm) $\downarrow$ & \textbf{21.45} & 62.33 & 30.80 & 24.92 \\
\midrule
Orientation error ($^\circ$) $\downarrow$ & 37.8 & 47.6 & 23.4 & 26.2 \\
Contact retention (\%) $\uparrow$ & 92 & 2 & 89 & 89 \\
Slip, sim / ref.\ (cm/s) & 27.7 / 19.4 & 20.0 / 26.4 & 20.0 / 20.8 & 19.6 / 20.5 \\
\midrule
Penetration (\% frames) $\downarrow$ & 0.0 & 0.0 & 0.0 & 0.0 \\
Object below ground (\% frames) $\downarrow$ & 0.02 & 0.0 & 0.0 & 0.0 \\
Joint-limit violations (\% frames) $\downarrow$ & 0.60 & 0.03 & 0.00 & 0.05 \\
Falls (\% clips) $\downarrow$ & 34 & 0 & 0 & 0 \\
\bottomrule
\end{tabular}
\end{table}

ULTRA is trained on the same training clips and split as our motor model. BFM-Zero and our motor model each receive the native reference prompt $\boldsymbol z_t^{\mathrm{ref}}$, the projected mean of $B$ over a short window of upcoming reference states, and the last row adds the drift correction of Sec.~\ref{sec:app:drift-correction} to the same checkpoint.

\subsection{Goal-conditioned task families}
\label{sec:app:tasks}

Table~\ref{tab:app:tasks} lists the eight families with the key constraint of each verifier, and Table~\ref{tab:app:task-results} reports success per family. Tasks with intermediate contacts or scene constraints are scored on the full verifier instead of terminal position error, so a missed contact mode stays visible.

\begin{table}[ht]
\centering
\small
\caption{\textbf{Task families for program search.} The verifier scores outcomes and constraints independently of the agent-written reward.}
\label{tab:app:tasks}
\begin{tabularx}{\linewidth}{@{}>{\raggedright\arraybackslash}p{0.25\linewidth}YY@{}}
\toprule
Task & Interaction & Key constraint \\
\midrule
Push to mark & Push object to a ground target & Keep it on the floor, stop near target \\
Carry at mid height & Acquire, raise, and carry & Hold a waist-relative height band \\
Carry at chest height & Acquire, raise, and carry & Hold a body-relative height \\
Tip onto a new face & Turn it over about a horizontal axis & Large tilt, left standing on the new face \\
Push through a gate & Push between two immovable walls & Hold a straight heading through the gap \\
Push around an obstacle & Push past a pillar in the way & Deviate, then recover the line \\
Kicking to mark & Move object with repeated foot contacts & No hand contact \\
Lift to support & Raise onto an elevated surface & Final height and placement \\
\bottomrule
\end{tabularx}
\end{table}

\noindent\textbf{Rare behaviors need focused reward inference.}
How the prompt of Eq.~\ref{eq:zr} weighs bank states matters most for rare target behaviors such as kicking. Uniform averaging washes out the few relevant states, whereas sharp focus rests the prompt on a handful of them, so success peaks at a moderate tilt $\beta$, and bank size shows the same trade-off (Table~\ref{tab:app:fb-bank}, Fig.~\ref{fig:app:fb-scaling}). Reward inference therefore needs \textit{enough relevant states} and a \textit{moderate focus}.
\begin{table}[ht]
\centering
\small
\caption{\textbf{Success per task family (\%).} CMA denotes CMA-ES. The macro-average weights families equally.}
\label{tab:app:task-results}
\setlength{\tabcolsep}{5pt}
\begin{tabular}{@{}lccc@{}}
\toprule
Task family & \shortstack{Human\\+ CMA} & \shortstack{Initial agent\\+ CMA} & InterEvolve \\
\midrule
Push to mark & 96.9 & \textbf{100.0} & \textbf{100.0} \\
Carry at mid height & 0.0 & 40.6 & \textbf{100.0} \\
Carry at chest height & 0.0 & 0.0 & \textbf{100.0} \\
Tip onto a new face & 0.0 & 0.0 & \textbf{95.3} \\
Push through a gate & 0.0 & 25.0 & \textbf{73.4} \\
Push around an obstacle & 0.0 & 4.7 & \textbf{71.9} \\
Kicking to mark & 17.2 & 37.5 & \textbf{59.4} \\
Lift to support & 29.7 & 68.8 & \textbf{92.2} \\
\midrule
Macro-average & 18.0 & 34.6 & \textbf{86.5} \\
\bottomrule
\end{tabular}
\end{table}

\begin{figure}[h]
  \centering
  \begin{minipage}[t]{0.495\linewidth}\centering
    \includegraphics[width=\linewidth]{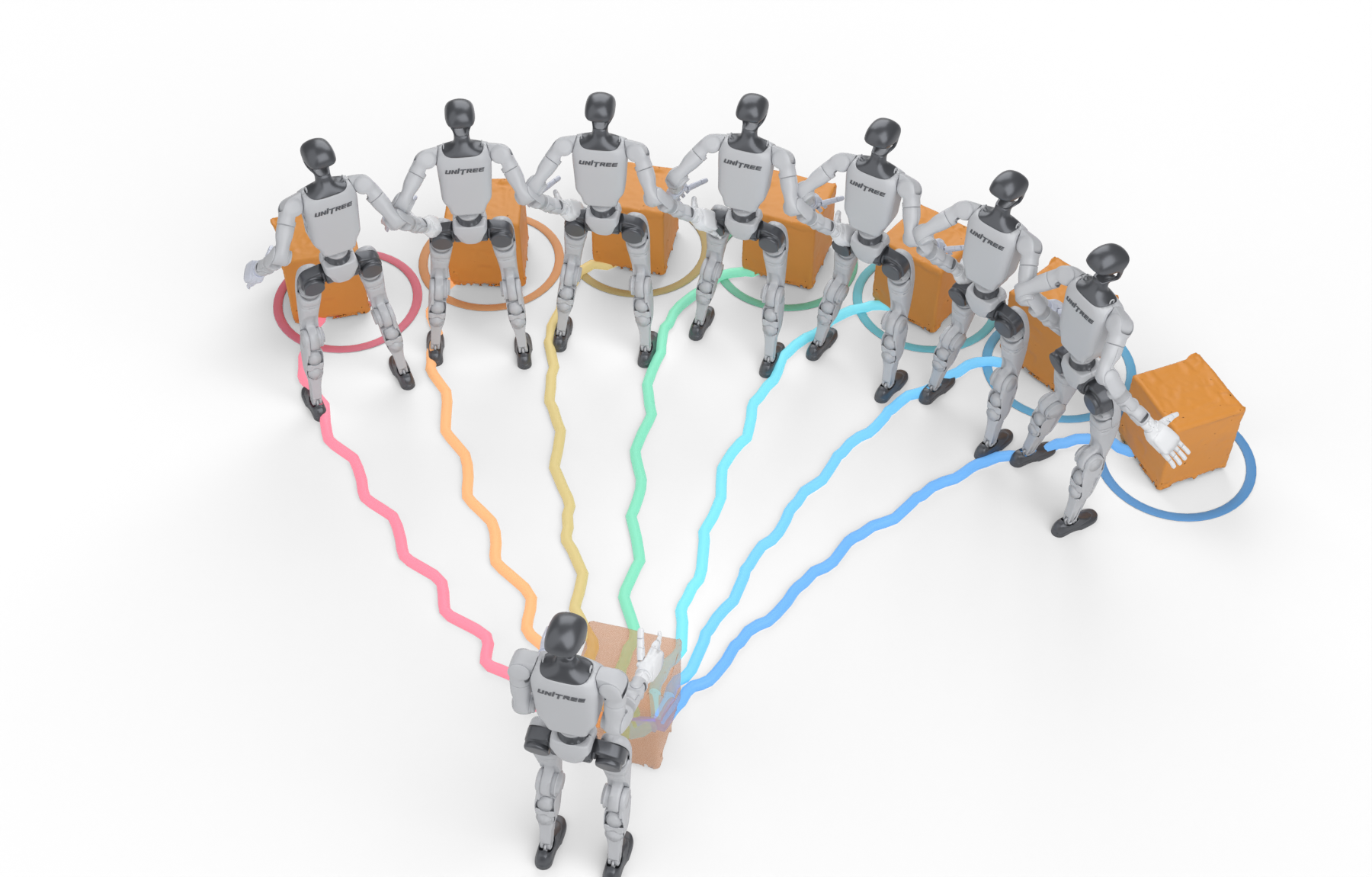}\\[-2pt]
    {\small (a) One start, different goals}
  \end{minipage}\hfill
  \begin{minipage}[t]{0.495\linewidth}\centering
    \includegraphics[width=\linewidth]{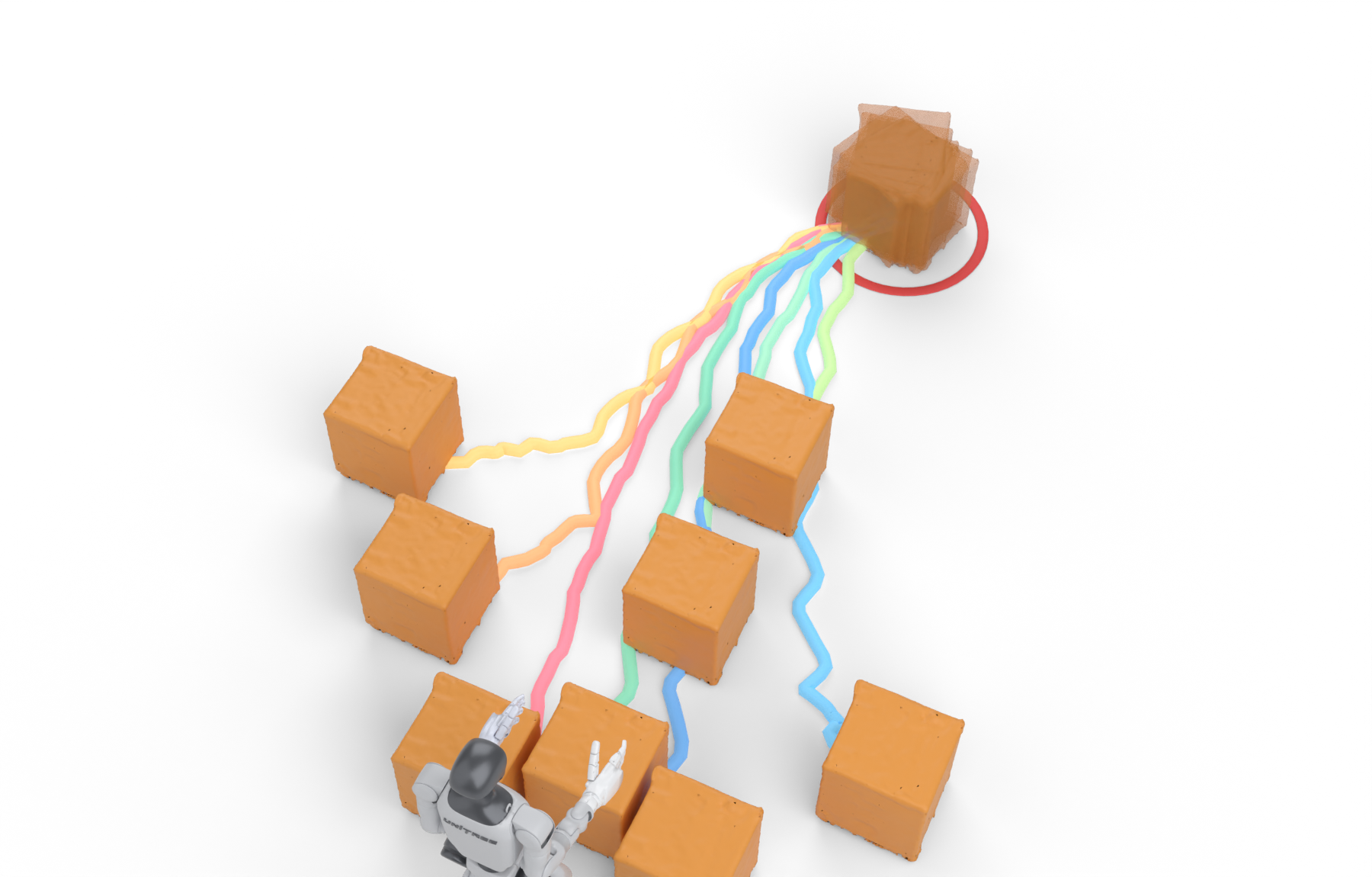}\\[-2pt]
    {\small (b) Different box starts, one goal}
  \end{minipage}
  \caption{{\textbf{Push to mark under new goals and starts}, with the selected program unchanged. (a) From one start, targets at bearings from $-45^\circ$ to $+45^\circ$, 2.5\,m away; rings mark the targets. (b) The box starts 0.55, 1.2, or 1.8\,m from the robot at bearings of $-40^\circ$, $0^\circ$, and $+40^\circ$, and is pushed to one target (red ring). Lines trace the box; each rollout shown is one of 16 per setting.}}
  \label{fig:app:variation}
\end{figure}

\noindent{\textbf{Varying the goal and the start.}
Fig.~\ref{fig:app:variation} runs the push-to-mark program selected in Sec.~\ref{sec:exp:skills}, with its stages and constants unchanged, in two settings it was not searched on but generalize well. Search placed the target 2.5\,m straight ahead of a box just in front of the robot. In (a) the target moves to other bearings, and in (b) the box starts at other distances and bearings around the robot while the target stays fixed. In both settings the controller walks to the box and pushes it to the target; success drops only at the widest angles.}

\noindent{\textbf{More objects.}
Table~\ref{tab:app:object-transfer} runs the eight selected programs, searched on the large box, unchanged on the plastic box, the small box, and the suitcase, which the controller saw in pretraining but no program saw during search. Rerunning the protocol on the large box reproduces Table~\ref{tab:app:task-results} exactly. Transfer depends on the family and the object: lifting onto a support keeps 70--92\% success on all three boxes, and the suitcase keeps most of the carrying and obstacle success, whereas push to mark reaches at most 25\% on the other boxes even with its height band shifted, and the small box is almost never pushed around the obstacle. Averaged over families, the unchanged programs reach 36--59\% on the new boxes. Adapting each program to the new box, raises the averages to 79--89\%, each adapted program confirmed once with the same protocol; kicking the plastic and small boxes and pushing the plastic box around the obstacle did not improve and keep the unchanged program.}

\begin{table}[h]
\centering
\small
\caption{{\textbf{Selected programs on other boxes.} Each program of Table~\ref{tab:app:task-results} is searched on the large box and run on the three other box-like OMOMO objects the controller was trained on, first unchanged (left of the arrow) and then adapted to each box by hand; $\dagger$ marks boxes where adaptation did not help and the unchanged program is kept. SR (\%) over 64 test scenarios, as in Table~\ref{tab:app:task-results}; the large-box column is a rerun of the same protocol and reproduces that table exactly.}}
\label{tab:app:object-transfer}
\setlength{\tabcolsep}{4pt}
\begin{tabular}{lcccc}
\toprule
{Task family} & {Large box} & {Plastic box} & {Small box} & {Suitcase} \\
\midrule
{Push to mark} & {100.0} & {0.0 (4.7) $\to$ 98.4} & {0.0 (0.0) $\to$ 98.4} & {0.0 (25.0) $\to$ 92.2} \\
{Carry at mid height} & {100.0} & {15.6 $\to$ 84.4} & {35.9 $\to$ 96.9} & {76.6 $\to$ 90.6} \\
{Carry at chest height} & {100.0} & {18.8 $\to$ 98.4} & {29.7 $\to$ 95.3} & {98.4$^\dagger$} \\
{Tip onto a new face} & {95.3} & {78.1 $\to$ 93.8} & {28.1 $\to$ 89.1} & {31.2 $\to$ 96.9} \\
{Push through a gate} & {73.4} & {40.6 $\to$ 98.4} & {56.2 $\to$ 85.9} & {68.8 $\to$ 90.6} \\
{Push around an obstacle} & {71.9} & {32.8$^\dagger$} & {1.6 $\to$ 84.4} & {96.9$^\dagger$} \\
{Kicking to mark} & {59.4} & {32.8$^\dagger$} & {65.6$^\dagger$} & {7.8 $\to$ 18.8} \\
{Lift to support} & {92.2} & {81.2 $\to$ 89.1} & {70.3 $\to$ 93.8} & {92.2$^\dagger$} \\
\midrule
{Macro-average} & {86.5} & {37.5 $\to$ 78.5} & {35.9 $\to$ 88.7} & {59.0 $\to$ 84.6} \\
\bottomrule
\end{tabular}
\end{table}

\subsection{Fixed-program baselines}
\label{sec:app:baselines}

Human and initial agent programs are each evaluated before and after CMA-ES with the same feature access, parameter bounds, and rollout budget per program. Because reward-tilted inference couples reward scale with $\beta$ (Sec.~\ref{sec:app:aggregation}), we hold $\beta$ fixed when comparing reward formulations.

\subsection{Search-design ablations}
\label{sec:app:search-ablations}

Table~\ref{tab:search:mechanisms} removes one of the five search components below at a time. Every variant runs on the same task families, controller, and verifier, and is scored on the program its selection rule returns.

\noindent\textbf{CMA-ES tuning.}
In the full system, the inner loop tunes the constants $\boldsymbol\theta$ of every valid proposal with CMA-ES before the proposal is compared with the current best program (Sec.~\ref{sec:app:compute}). Without it, each proposal runs with the initial constants the agent wrote. This isolates how much of the gain comes from calibrating a given structure.

\noindent\textbf{Targeted edits.}
In the full system, the agent sees the current best program and its feedback and is instructed to repair it: to change the stage or term that caused the failing criterion, or to merge stages from programs that lead on different criteria (Sec.~\ref{sec:app:context}). The full-rewrite variant keeps the same context, feedback, acceptance rule, and budget, but asks for new programs written from scratch each round. This isolates the value of editing a working program over regenerating it.

\noindent\textbf{Multi-scenario evaluation.}
In the full system, every candidate is evaluated on a grid of multiple search scenarios that vary the object's bearing and distance (Sec.~\ref{sec:app:selection}). The single-scenario variant evaluates each candidate on one scenario and spends the saved budget on more candidates. This isolates whether judging a candidate across varied starts helps select edits that generalize rather than edits that happen to work once.

\noindent\textbf{Scene context.}
In the full system, the prompt describes the scene, such as object placement, supports, and obstacles, next to the task text. The variant gives the agent only the task text. The controller still observes the full scene, so this ablation affects planning but leaves feedback control intact.

\noindent\textbf{Multi-stage programs.}
In the full system, a program may split the task into up to three stages with their own rewards and completion conditions. The single-stage variant must express the whole task with one reward from the same feature vocabulary under the same budget. This isolates the value of temporal composition.

\subsection{Skill-library reuse}
\label{sec:app:library}

Library experiments fix everything except the library content, including the controller, the bank, and the target-task budget. The source library is frozen before target search and excludes the target task instances, so target rollouts inform only within-task revision. Table~\ref{tab:search:library} varies library content on composite tasks.

\noindent\textbf{Composite tasks.}
Table~\ref{tab:app:composite-tasks} describes the three composite tasks of Table~\ref{tab:search:library}. Each combines several contact modes that appear separately among the eight task families of Table~\ref{tab:app:tasks}, in one episode and from one initial reset: carrying and placing on a support, lifting and lowering onto another object, and pushing followed by kicking. As in the other families, the verifier scores the outcome and its constraints independently of the agent-written reward, and success requires every criterion in the same rollout. None of the three tasks, and no program written for them, is in any source library.

\begin{table}[ht]
\centering
\small
\caption{\textbf{Composite tasks for library reuse.} Distances are measured from the box's starting position.}
\label{tab:app:composite-tasks}
\begin{tabularx}{\linewidth}{@{}>{\raggedright\arraybackslash}p{0.18\linewidth}YY@{}}
\toprule
Task & Scene and interaction & Verifier criteria \\
\midrule
Relocate between supports & The box starts on a 0.3\,m support in front of the robot. The robot lifts it, turns around, carries it 2.9\,m to a 0.6\,m support behind it, sets it down, and lets go. & Stays upright; holds the box for a large part of the episode; lifts it above the destination height; ends within 0.3\,m of the mark on the support; hands clear of the box at the end. \\
Stack two boxes & Box A is on the floor 0.5\,m ahead; box B, of the same size and free to move, is 1.9\,m to the left. The robot lifts A, carries it above B, lowers it onto B's top face, and lets go. & Stays upright; lifts A clear of B's top; sustained two-handed contact; A stays level; ends within 0.25\,m of B's top (the target follows B if B is pushed); hands clear; A at rest. \\
Carry, place, and kick & The box is on the floor 0.5\,m ahead. The robot pushes it about 1.0\,m along the ground with its hands, lets go, and kicks it with its feet a further 1.5\,m to a mark 3.0\,m ahead. & Stays upright; box stays low (no lift); sustained hand contact during the push; at least one foot contact; box travels at least 1.0\,m after the last hand contact; hands clear; ends within 0.3\,m of the mark. \\
\bottomrule
\end{tabularx}
\end{table}

\noindent\textbf{Whether programs or lessons carry over.}
Table~\ref{tab:app:program-library} extends the carry, place, and kick task of Table~\ref{tab:search:library} with more library variants, all scored on the same 64 test scenarios. Lessons are the library's shared design notes without its programs summarized by human. Run as stored, without search, only the kicking program solves any scenario (6.3\%), so the gain comes from search adapting stored programs rather than from copying one. After three search rounds, the full library reaches 60.9\%, programs without lessons 48.4\%, lessons without programs 21.9\%, and neither 12.5\%. Among program subsets, the three carrying programs reach 23.4\% and every subset that contains the kicking program reaches 43.8--53.1\%, all below the full library.

\begin{table}[ht]
\centering
\small
\caption{\textbf{Library ablations on carry, place, and kick}, extending Table~\ref{tab:search:library}. Every row is scored on the same 64 test scenarios, none of which any search used; SR is the percentage of scenarios solved, where a scenario is solved when one of three rollouts meets every criterion. Left: each stored program run as written, without search. Right: three search rounds from each source library; the shaded row is the full library of Table~\ref{tab:search:library}.}
\label{tab:app:program-library}
\setlength{\tabcolsep}{4pt}
\begin{minipage}[t]{0.37\linewidth}
\centering
\begin{tabular}[t]{@{}lc@{}}
\toprule
Stored program, no search & SR (\%) \\
\midrule
Kicking to mark & 6.3 \\
Push to mark & 0.0 \\
Carry at mid height & 0.0 \\
Carry at chest height & 0.0 \\
Push around an obstacle & 0.0 \\
Push through a gate & 0.0 \\
Lift to support & 0.0 \\
Tip onto a new face & 0.0 \\
\bottomrule
\end{tabular}
\end{minipage}\hfill
\begin{minipage}[t]{0.60\linewidth}
\centering
\begin{tabular}[t]{@{}lccc@{}}
\toprule
Source library & Programs & Lessons & SR (\%) \\
\midrule
\rowcolor{gray!10}
Full & 8 & \checkmark & \textbf{60.9} \\
Programs only & 8 & -- & 48.4 \\
Lessons only & 0 & \checkmark & 21.9 \\
None & 0 & -- & 12.5 \\
\midrule
Tip, lift, chest carry, kick & 4 & \checkmark & 50.0 \\
Lift, kick & 2 & \checkmark & 53.1 \\
Kick only & 1 & \checkmark & 43.8 \\
Carry only (three carrying) & 3 & -- & 23.4 \\
\bottomrule
\end{tabular}
\end{minipage}
\end{table}

\subsection{Coverage of the pretraining references}
\label{sec:app:coverage}

\begin{figure}[t]
  \centering
  \includegraphics[width=0.72\linewidth]{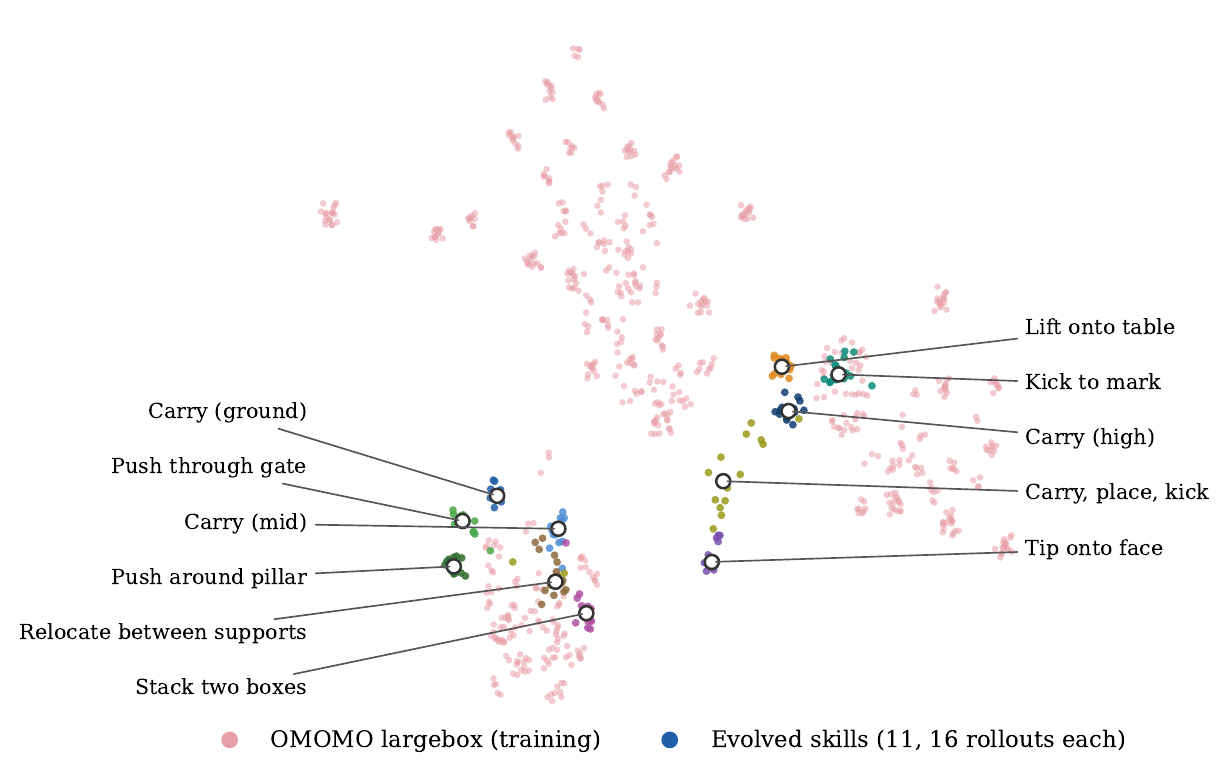}
  \caption{\textbf{Evolved skills versus pretraining references.} Each point is one motion: an OMOMO largebox training clip (pink) or one rollout of an evolved skill (colored, 16 per skill; circles mark skill centers). A motion is described by its body posture relative to the pelvis and heading, summarized over time, and embedded with UMAP.}
  \label{fig:umap}
\end{figure}

Fig.~\ref{fig:umap} compares evolved behaviors with the data the FB model was pretrained on. Carrying at ground height and pushing stay inside the reference cloud, whereas lifting onto a table, carrying at chest height, kicking, tipping, and carrying, placing, then kicking form their own clusters, so the InterEvolve can produce motions that no training reference contains. 

\textbf{Motions.} We compare 729 OMOMO largebox clips and the evolved 11 skills with 16 rollouts each, cut to the interaction window (from 0.5\,s before the object first moves to 0.5\,s after it last moves). 

\textbf{Descriptor.} Only the body is used, so object size and scene layout cannot separate the two sets. For each frame, the 29 link positions are taken relative to the pelvis and rotated into the robot's own heading, so walking direction and turning do not count as different motions. Each coordinate is resampled to 128 steps and summarized by its first 16 DCT coefficients, a low-frequency description of how the posture changes over the window. 

\textbf{Embedding.} Features are standardized with the reference statistics, reduced by PCA fit on the references (32 components, 91\% of variance), and embedded jointly with UMAP. 

\textbf{Distance check.} Independently of the 2D layout, we measure each rollout's distance to its nearest reference clip in the PCA space and rank it against the distance of each reference clip to its nearest reference from a different recording. The median rollout ranks at the 76th percentile; lifting onto a table, carrying at chest height, and carrying, placing, then kicking rank at the 98th, and carrying at ground height and pushing around a pillar at the 18th and 9th.

\subsection{Transfer and physical deployment}
\label{sec:app:system-evaluation}

\noindent\textbf{Perception.}
We deploy on a Unitree G1 whose policy commands the same 29 joints as in simulation at 50\,Hz. An RGB-D camera on the torso streams color and depth to an off-board GPU workstation, which runs perception, state estimation, and the policy. FastSAM~\citep{zhao2023fastsam} and CLIP~\citep{radford2021clip} find the box in the image, and FoundationPose~\citep{wen2024foundationpose} registers a mesh of the box on this mask and tracks its 6-D pose frame by frame, discarding poses that disagree with the measured depth. Each pose is time-stamped at capture and mapped into the robot frame with the joint angles and leg odometry of that moment. A filter fuses these poses with box fits from a LiDAR on the torso and with floor and contact constraints, and uses the camera only while the box is free, since the arms occlude it during manipulation. From the filtered pose, the controller computes the same object features it observes in simulation.

\noindent\textbf{Execution.}
A program is evolved and verified in simulation, and the latent prompts it produces during a simulated rollout are replayed on the robot, one per control step, while the frozen controller closes the loop on proprioception and the estimated object pose. Execution starts once the object estimate is confident, and it pauses while the estimate is invalid.

\section{Limitations} Program search is bounded by the motor repertoire, the reward-inference bank, and the available measurements. It cannot elicit behavior the controller never learned. Each search also costs GPU-hours of simulation and LLM latency, which rules out real-time replanning during physical execution. Letting accumulated experience also update the controller is a next step.

\end{document}